\documentclass[twoside,11pt]{article}

\usepackage{jmlr2e}

\usepackage{blkarray}
\usepackage{csquotes}
\usepackage{soul}
\usepackage{amsmath}
\usepackage{mathtools}
\usepackage{bbm}
\usepackage{verbatim}
\usepackage{hyperref}

\usepackage[capitalize]{cleveref}
\crefformat{equation}{(#2#1#3)}
\crefrangeformat{equation}{(#3#1#4) to~(#5#2#6)}
\crefmultiformat{equation}{(#2#1#3)}%
{ and~(#2#1#3)}{, (#2#1#3)}{ and~(#2#1#3)}
\makeatletter
\if@cref@capitalise
\crefname{proposition}{Proposition}{Propositions}
\crefname{definition}{Definition}{Definitions}
\else
\crefname{proposition}{proposition}{propositions}
\crefname{proposition}{definition}{definitions}
\fi
\makeatother
\usepackage{autonum}

\newcommand{\R}{\mathbb{R}}
\newcommand{\G}{\mathcal{G}}
\newcommand{\A}{\mathcal{A}}

\newcommand{\M}{\mathcal{M}}

\newcommand\independent{\protect\mathpalette{\protect\independenT}{\perp}}
\def\independenT#1#2{\mathrel{\rlap{$#1#2$}\mkern2mu{#1#2}}}
\newcommand{\PD}{\R_{\text{spsd}}^{n\times n}}
\newcommand{\Gd}{f_{\mu,\Sigma}}

\DeclareMathOperator{\sgn}{sgn}
\DeclareMathOperator{\CI}{CI}
\DeclareMathOperator{\pa}{pa}
\DeclareMathOperator{\diag}{diag}
\DeclareMathOperator{\KL}{KL}
\DeclareMathOperator{\tr}{tr}
\DeclareMathOperator{\F}{F}

\usepackage{tikz}
\newcommand{\xx}{1}
\newcommand{\yy}{1}

\jmlrheading{1}{2017}{1-20}{4/00}{}{GL17}{Christiane G\"{o}rgen and Manuele Leonelli}

\ShortHeadings{Model-Preserving Sensitivity Analysis}{G\"{o}rgen and Leonelli}
\firstpageno{1}

\begin{document}

\title{Model-Preserving Sensitivity Analysis for Families of Gaussian Distributions}

\author{\name Christiane G\"{o}rgen \email christiane.goergen@mis.mpg.de \\
       \addr Max Planck Institute for Mathematics in the Sciences\\
       Inselstra{\ss}e 22, \\
       04103 Leipzig, Germany 
       \AND
       \name Manuele Leonelli \email m.leonelli@glasgow.ac.uk \\
       \addr School of Mathematics and Statistics\\
       University of Glasgow\\
       Glasgow G12 8SQ, UK}

\editor{Kevin Murphy and Bernhard Sch{\"o}lkopf}

\maketitle

\begin{abstract}
The accuracy of probability distributions inferred using machine-learning algorithms heavily depends on data availability and quality. In practical applications it is therefore fundamental to investigate the robustness of a statistical model to misspecification of some of its underlying probabilities. In the context of graphical models, investigations of robustness fall under the notion of sensitivity analyses. These analyses consist in varying some of the model's probabilities or parameters and then assessing how far apart the original and the varied distributions are.  However, for Gaussian graphical models, such variations usually make the original graph an incoherent  representation of the model's conditional independence structure. Here we develop an approach to sensitivity analysis which guarantees the original graph remains valid after any probability variation and we quantify the effect of such variations using different measures. To achieve this we take advantage of algebraic techniques to both concisely represent conditional independence and to provide a straightforward way of checking the validity of such relationships. Our methods are demonstrated to be robust and comparable to standard ones, which break the conditional independence structure of the model, using an artificial example and a medical real-world application.
\end{abstract}

\section{Introduction}
The validation of both machine-learnt and expert-elicited statistical models is one of the most critical phases of any applied analysis. Broadly speaking, this validation phase consists of checking that a model produces outputs that are in line with current understanding, following a defensible and expected mechanism \citep{French2003,Pitchforth2013}. A critical aspect of such a validation is the investigation of the effects of variations in the model's inputs to outputs of interest. These types of investigations are usually referred to as \textit{sensitivity analyses} \citep{Borgonovo2016}.

Various sensitivity methods are now in place for generic statistical models \citep{Saltelli2000}. A large proportion of these have focused on graphical models \citep{Lauritzen1996} and, in particular, on Bayesian network (BN) models \citep{Pearl1988}. A BN is a graphical representation of a statistical model defined via a set of conditional independence statements \citep{Dawid1979}.

Sensitivity analysis in BNs usually consists of two main steps. First \textit{local} changes on outputs of interest are investigated via sensitivity functions: so probabilities are studied as functions of the input parameters as these vary in some appropriate interval. Once possible input parameter changes have been identified, the \textit{global} effects that these would have on the overall distribution of the network are studied. These global effects are usually quantified by some divergence or distance between the original and the varied distributions, for instance using the Kullback-Leibler (KL) divergence \citep{Kullback1951}.

Sensitivity methods in BNs usually focus on either models consisting of discrete random variables or, in the continuous case, of multivariate Gaussian distributions. The properties of sensitivity functions in the discrete case have been studied extensively \citep{Castillo1997,Coupe2002}. Here, when some probabilities of interest are varied, then some others, namely those associated to the same conditional probability tables, need to \emph{covary} in order to respect the sum-to-one condition of probabilities. The gold-standard in this context is \emph{proportional covariation} which assigns to the covarying parameters the same proportion of the remaining probability mass as they originally had \citep{Renooij2014}. The use of proportional covariation is justified by a variety of optimality criteria because it often minimizes the distance between the original and varied distribution amongst all possible covariations \citep{Chan2002, Leonelli2017}. 
In the continuous case, the properties of sensitivity functions for Gaussian BNs have been known for quite some time \citep{Castillo1997}. They are rational functions of both the mean parameters and the entries of the covariance matrix of the Gaussian distribution associated to the network. Following these early developments, methods to quantify the distance between the original distribution and the one obtained from perturbations of the mean vector and covariance matrix were introduced \citep{Gomez2007,Gomez2008, Gomez2013}, entailing the computation of the KL divergence between the two distributions. 

One of the major drawbacks of the established Gaussian sensitivity methods is that in most cases perturbations of the covariance matrix make the graph of the original BN a non-faithful representation of the new distribution. This is because entries of the covariance matrix relate directly to conditional independence relationships between the depicted variables. In the discrete case this issue does not arise since a perturbation is applied directly to the conditional probability distributions associated to the BN rather than to the covariance structure of the model so that any new distribution automatically respects all the conditional independences of the model.

In practice, however,  Gaussian BN users may want to apply a perturbation to some parameters whilst retaining the original graphical structure of their model and all of its entailed conditional independences. To tackle this issue we introduce a new class of perturbations of Gaussian vectors, called \emph{model-preserving}, which have the property that the graphical representation of the original distribution remains valid after the perturbation. Whilst standard sensitivity methods act additively over the entries of the covariance matrix of  the underlying Gaussian distribution, our model-preserving approach acts multiplicatively as formalized below.  Furthermore, and conversely to standard sensitivity methods which only vary the entries of interest of the covariance matrix, in model-preserving perturbations additional parameters need to covary so that all conditional independences of the model are retained.  In particular, this covariation ensures that the matrix under perturbation remains a covariance matrix of the original Gaussian model. This can be thought of as the continuous analogue of covariation techniques in the discrete case in the sense that it ensures that the varied object remains inside its original class.

Below we introduce various ways to select the parameters that need to covary for a given perturbation and we quantify the distance between the original and varied distributions using a variety of measures. We achieve this by adopting an algebraic approach which characterizes conditional independences as specific vanishing minors of a covariance matrix. Algebraic methods have been already used extensively in machine learning problems \citep[see for instance][]{Rusakov2005,Zwiernik2011} but, to the best of our knowledge, we provide here their first application to sensitivity studies.

\section{Conditional Independence and Gaussian Graphical Models}\label{sec:background}
We start by reviewing the theory of Gaussian conditional independence models. We then focus on two graphical representations of specific sets of conditional independences, namely undirected and directed graphical models.

\subsection{Gaussian Conditional Independence Models}
Let $Y$ be a $n$-dimensional Gaussian random vector with mean $\mu\in\mathbb{R}^n$ and covariance matrix $\Sigma\in\PD$, where $\PD\subset\mathbb{R}^{n\times n}$ denotes the cone of symmetric, positive semidefinite $n\times n$ matrices. Let $\Gd$  be the density of a Gaussian distribution parametrized by $\mu$ and $\Sigma$. For index sets $A,B\subseteq[n]=\{1,\dots,n\}$, let $\mu_A=(\mu_i)_{i\in A}$ be the subvector of the mean with entries indexed by $A$ and  $\Sigma_{A,B}$ be the submatrix of $\Sigma$ with rows indexed by $A$ and columns indexed by $B$. Both marginal and conditional distributions of Gaussian vectors are Gaussian. In particular, for any two disjoint sets $A,B\subset [n]$, the random vector $Y_A=(Y_i)_{i\in A}$ has density $f_{\mu_A,\Sigma_{A,A}}$ and $Y_A|Y_B=y_B$ has density $f_{\mu^{A|B}, \Sigma^{A|B}}$ where 
\begin{equation}
\label{eq:moments}
\mu^{A|B}=\mu_A+\Sigma_{A,B}\Sigma_{B,B}^{-1}(y_B-\mu_B)  \mbox{ and } \Sigma^{A|B}=\Sigma_{A,A}-\Sigma_{A,B}\Sigma_{B,B}^{-1}\Sigma_{B,A}.
\end{equation}

In this paper we consider Gaussian models defined by sets of conditional independence statements. The random vector $Y_A$ is henceforth said to be \emph{conditionally independent of} $Y_B$ \emph{given} $Y_C$ for disjoint subsets $A,B,C\subseteq[n]$ if and only if the density factorizes as
\begin{equation}
f_{\mu^{A\cup B|C},\Sigma^{A\cup B|C}}=f_{\mu^{A|C},\Sigma^{A|C}}f_{\mu^{B|C},\Sigma^{B|C}}.
\end{equation}
We sometimes abbreviate this statement to $A\independent B~|~C$. The following proposition from \citet{Drton2008} demonstrates that conditional independence relationships in multivariate Gaussian models can be characterized in a straightforward algebraic way. 

\begin{lemma}[Proposition 3.1.13 of \citet{Drton2008}]
\label{prop:drton}
For a $n$-dimensional Gaussian random vector $Y$ with density $\Gd$ and disjoint $A,B,C\subset[n]$, the conditional independence statement $A\independent B~|~C$ is true if and only if all $(\#C +1)\times (\#C +1)$ minors of the matrix $\Sigma_{A\cup C,B\cup C}$ are equal to zero. Here, $\#C$ denotes the cardinality of the set $C$.
\end{lemma}

This duality between conditional independence and the vanishing of a set of equations provides the key insight on which we build our new algebraic sensitivity methods. In particular, in the subsequent sections we easily establish techniques for Gaussian graphical models which ensure that if a set of equations vanished before a perturbation of one or multiple entries of the covariance matrix then it will continue to vanish after an appropriate covariation of some of the other entries of that matrix.

Let henceforth $\CI=\{A_1\independent B_1~|~C_1,\dots, A_r\independent B_r~|~C_r\}$  denote a set of conditional independence statements for disjoint index sets $A_i,B_i,C_i\subset[n]$ and $i\in[r]$, with $r\in\mathbb{N}$. A \textit{Gaussian conditional independence model} $\mathcal{M}_{\CI}$ for a $n$-dimensional random vector $Y$ for which all $\CI$ statements are true is a special subset of all possible Gaussian densities $\Gd$:
\begin{equation}\label{eq:cimodel}
\M_{\CI}\subseteq\{\Gd~|~\mu\in\mathbb{R}^n,\Sigma\in\PD\}.
\end{equation}
By Lemma~\ref{prop:drton}, the parameter space of $\mathcal{M}_{\CI}$ is equal to the algebraic set 
\begin{equation}\label{eq:parmset}
\begin{split}
\A_{\CI}=\{\mu\in\mathbb{R}^n,\Sigma\in\PD~|~&g(\Sigma)=0\text{ for all polynomials } g \text{ which are }\\
&(\#C_i+1)\times(\#C_i+1) \text{ minors of }\Sigma_{A_i\cup C_i, B_i\cup C_i}, i\in[r]\}.
\end{split}
\end{equation}
Thus, every Gaussian conditional independence model is the image of a bijective parametrization map $(\mu,\Sigma)\mapsto\Gd$ whose domain is given by equation (\ref{eq:parmset}).

\begin{example}
\label{ex:first}
Let $Y_1$, $Y_2$ and $Y_3$ be jointly Gaussian and suppose $Y_3\independent Y_1~|~Y_2$. Then by Proposition~\ref{prop:drton}, the $2\times 2$ minors of the submatrix 
\begin{equation}\label{eq:matrix}
\Sigma_{\{2,3\},\{1,2\}}=
\begin{pmatrix}
\sigma_{21}&\sigma_{22}\\
\sigma_{31}& \sigma_{32}
\end{pmatrix}
\end{equation}
need to vanish. Here the only vanishing minor simply corresponds to the determinant. So $g=\sigma_{21}\sigma_{32}-\sigma_{31}\sigma_{22}$ is a polynomial which is zero in equation (\ref{eq:parmset}).
\end{example}

\begin{example}
\label{ex:second}
For a Gaussian random vector $Y=(Y_i)_{i\in [4]}$ together with the conditional independence $Y_2\independent \{Y_1,Y_3\}~|~ Y_4$, the $2\times 2$ minors of the submatrix
\begin{equation}
\Sigma_{\{2,4\},\{1,3,4\}}=
\begin{pmatrix}
\sigma_{21}&\sigma_{23}&\sigma_{24}\\
\sigma_{41}&\sigma_{43}&\sigma_{44}
\end{pmatrix}
\end{equation}
need to vanish. Explicitly, $\sigma_{21}\sigma_{43}-\sigma_{41}\sigma_{23}=0$, $\sigma_{21}\sigma_{44}-\sigma_{41}\sigma_{24}=0$ and $\sigma_{23}\sigma_{44}-\sigma_{43}\sigma_{24}=0$.
\end{example}

The following two sections review some basic results on directed and undirected graphical models. In particular, here we recall a second duality: the one between graphs and conditional independence relationships. In conjunction with Lemma~\ref{prop:drton}, these form the basis for algebraic sensitivity methods which ensure that after covariation a graph remains a faithful representation of the model.

\subsection{Undirected Gaussian Graphical Models}
For Gaussian random vectors, many sets of conditional independences can be represented visually by a graph. We start by defining families of Gaussians supported by undirected graphs.

\begin{definition}\label{def:undirected}
A \emph{Gaussian undirected graphical model} for a random vector $Y=(Y_i)_{i\in[n]}$ is defined by an undirected graph $\G=(V,E)$ with vertex set $V=[n]$ and a family of densities $\Gd$ whose covariance matrix $\Sigma$ is such that $(\Sigma^{-1})_{ij}=0$ if and only if $(i,j)\not\in E$.
\end{definition}

Thus the zero entries in the inverse of the covariance matrix of a Gaussian undirected graphical model correspond to conditional independence statements. This is  usually called the \emph{pairwise Markov} property \citep{Lauritzen1996}. In particular, if $(\Sigma^{-1})_{ij}=0$ then ${Y_i\independent Y_j~|~ Y_{[n]\setminus \{i,j\}}}$: so the absence of an edge between two random variables implies that these are conditionally independent given all the others.

By Lemma \ref{prop:drton} and Definition~\ref{def:undirected}, the fact that an entry in the inverse of the covariance matrix is equal to zero exactly corresponds to the vanishing of the minors of an appropriate submatrix of $\Sigma$.

\begin{figure}
\centering
\begin{tikzpicture}
\renewcommand{\yy}{0.8}
\node (x1) at (0,0) {1};
\node (x2) at (\xx,\yy) {2};
\node (x3) at (\xx,-\yy) {3};
\node (x4) at (2*\xx,0) {4};
\draw (x2) -- (x4) -- (x3) -- (x1) -- (x4);
\end{tikzpicture}
\caption{An undirected graph for the conditional independence model ${Y_2\independent \{Y_1,Y_3\}~|~ Y_4}$ in Examples~\ref{ex:second} and~\ref{ex:UG}. \label{fig:UG}}
\end{figure}
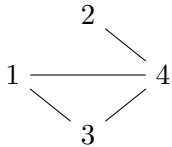

\begin{example}[Example \ref{ex:second} continued]
\label{ex:UG}
The statement $Y_2\independent \{Y_1,Y_3\}~|~ Y_4$ can be represented by the undirected graph in  Figure~\ref{fig:UG} where the edges $(1,2)$ and $(2,3)$ are not present. 
\end{example}

\subsection{Gaussian Bayesian Networks}
\label{sec:BN}
For directed graphical models, conditional independence relationships cannot be explicitly represented by zeros in the inverse of the covariance matrix. Gaussian BNs can however be constructed from the definition of a conditional univariate Gaussian distribution at each of its vertices \citep{Richardson2002}. 

\begin{definition}
\label{def:GBN}
A \emph{Gaussian Bayesian network} for a random vector $Y=(Y_i)_{i\in[n]}$ is defined by a directed acyclic graph $\G=(V,E)$  with vertex set $V=[n]$ such that to each $i\in[n]$ is associated a conditional Gaussian density $f_{\mu_i,\sigma_i}$ with mean $\mu_i=\beta_{0i}+\sum_{j\in\pa(i)}\beta_{ji}y_j$ and variance $\sigma_{i}\in\mathbb{R}_{+}$. Here, $\pa(i)\subseteq[i-1]$ denotes the parent set of the vertex $i$ in $\G$, $\beta_{0i}\in\R$ and $\beta_{ji}\in\R$ for all $j\in\pa(i)$.
\end{definition}

The Gaussian densities $f_{\mu_i,\sigma_i}$ in a Gaussian BN are associated to conditional independence statements of the form ${Y_i\independent Y_{[i-1]\setminus \pa(i)}~|~Y_{\pa(i)}}$. Definition~\ref{def:GBN} then assigns a multivariate Gaussian distribution to the full vector $Y$ as follows. Let $\beta_0=(\beta_{0i})_{i\in[n]}$,   $B$ be the strictly upper triangular matrix with entries $B_{ji}=\beta_{ji}$ if $j\in\pa(i)$ and zero otherwise, and  $\Phi=\diag(\sigma_1,\dots,\sigma_n)$ be the diagonal matrix of the conditional variances. Then $Y$ has Gaussian density $\Gd$ with mean $\mu=(I-B)^{-\top}\beta_0$ and covariance matrix $\Sigma=(I-B)^{-\top}\Phi (I-B)^{-1}$ where $I$ denotes the identity. A matrix $\Sigma$ constructed in this way naturally respects the conditional independences associated to a directed acyclic graph $\G$. 
In general, whether or not a covariance matrix can be associated to a specific directed acyclic graph can be checked by the vanishing of the minors of appropriate submatrices of~$\Sigma$, as formalized in Proposition~\ref{prop:drton}.

\begin{example}[Example \ref{ex:first} continued]
\label{ex:BN}
The conditional independence model ${Y_3\independent Y_1~|~Y_2}$ on three Gaussian random variables can be associated to the directed acyclic graph reported in Figure \ref{fig:BN} where the directed edge $(1,3)$ is not present. 
\end{example}

\begin{figure}
\centering
\begin{tikzpicture}
\renewcommand{\xx}{1.5}
\node (x1) at (0,0) {1};
\node (x2) at (\xx,0) {2};
\node (x3) at (2*\xx,0) {3};
\draw[->] (x1) -- (x2);
\draw[->] (x2) -- (x3);
\end{tikzpicture}
\caption{A directed acyclic graph for the conditional independence model ${Y_3\independent Y_1~|~Y_2}$ in Examples~\ref{ex:first} and~\ref{ex:BN}. \label{fig:BN}}
\end{figure}

\section{Sensitivity Analysis}\label{sect:reviewGBN}

We  now briefly review standard sensitivity methods for Gaussian BNs before we introduce our new model-preserving formalism. 

\subsection{Standard Methods for Gaussian Bayesian Networks}
\label{sec:stan}

Sensitivity methods for Gaussian BNs have been extensively studied \citep{Gomez2007,Gomez2008, Gomez2013}. For a generic Gaussian random vector $Y$ with density $f_{\mu,\Sigma}$, robustness is usually studied by perturbing the mean vector $\mu$ and the covariance matrix $\Sigma$. Such a perturbation is carried out by defining a perturbation vector $d\in\R^n$ and a matrix $D\in\R^{n\times n}$ which act additively on the original mean and variance, giving rise to a vector $\tilde{Y}$ with a new distribution$f_{\mu+d,\Sigma+D}$. 
The dissimilarity between these two vectors is then usually quantified via the KL divergence. 

For any two $n$-dimensional Gaussian vectors $Y$ and $Y'$ with distributions $f_{\mu,\Sigma}$ and $f_{\mu',\Sigma'}$ respectively, the \emph{KL divergence} between $Y$ and $Y'$ is given by
\begin{equation}
\label{eq:KL}
\KL(Y'||Y)=\frac{1}{2}\left(\tr(\Sigma^{-1}\Sigma')+(\mu-\mu')^{\top}\Sigma^{-1}(\mu-\mu')-n+\ln\left(\frac{\det(\Sigma)}{\det(\Sigma')}\right)\right).
\end{equation}
The KL divergence is not a distance and in particular violates the symmetry requirement, so in general $\KL(Y'||Y)\neq \KL(Y||Y')$. Symmetric extensions of KL divergences have been recently considered in sensitivity studies for Gaussian BNs \citep{Zhu2017} but a comprehensive review of these goes beyond the scope of this paper.

From equation (\ref{eq:KL}), the KL divergence between a perturbation $\tilde{Y}$ and the original $Y$ is
\begin{equation}\label{eq:KLknown}
\KL(\tilde{Y}||Y)=\frac{1}{2}\left(\tr(\Sigma^{-1}D)+d^{\top}\Sigma^{-1}d+\ln\left(\frac{\det(\Sigma)}{\det(\Sigma+D)}\right)\right).
\end{equation}
A proof of this statement is provided in the appendix.

In sensitivity analyses, the Gaussian vector $Y$ is often partitioned into two subvectors $Y_{E}$ and $Y_O$ such that $E\cup O=[n]$, including the evidential and output variables, respectively. Evidential variables are those for which a value $y_E$ is observed, whilst the output variables are those of interest to the user. Then also the perturbation mean vector and covariance matrix can be partitioned as
\begin{equation}
d=
\begin{pmatrix}
d_O\\
d_E
\end{pmatrix}
\quad\text{and}\quad
D=\begin{pmatrix}
D_{O,O} & D_{O,E}\\
D_{E,O} & D_{E,E}
\end{pmatrix}
\end{equation} 
where $D_{O,E}=D_{E,O}^{\top}$. This formalism enables the user to study the dissimilarity between the Gaussian vector  $Y_O|Y_E=y_E$ and perturbed output variables $\tilde{Y}_O|Y_E=y_E$ with distributions $f_{\mu^{O|E},\Sigma^{O|E}}$ and $f_{\tilde{\mu}^{O|E},\tilde{\Sigma}^{O|E}}$, respectively, where $\mu^{O|E}$ and $\Sigma^{O|E}$ are as in equation~(\ref{eq:moments}) and
\begin{align}
\tilde{\mu}^{O|E}&=\mu_O+d_O+(\Sigma_{O,E}+D_{O,E})(\Sigma_{E,E}+D_{E,E})^{-1}(y_E-\mu_E-d_E), \\
\tilde{\Sigma}^{O|E}&=\Sigma_{O,O}+D_{O,O}-(\Sigma_{O,E}+D_{O,E})(\Sigma_{E,E}+D_{E,E})^{-1}(\Sigma_{E,O}+D_{E,O}).
\end{align}

Efficient algorithms to propagate evidence $Y_E=y_E$ and to speedily compute these conditional distributions are available for Gaussian BNs \citep{Castillo2003,Malioutov2006}. The form of the KL divergence between these two conditional distributions depends on the block of parameters that are perturbed \citep[see][for more details]{Gomez2013}.

This standard approach has the critical drawback that if the Gaussian distribution is associated to a specific conditional independence model, for instance represented by a directed graph, then a perturbation may break its conditional independences. We illustrate this point below. 

\begin{example}[Example \ref{ex:BN} continued]
\label{ex:classicpert}
Suppose in the Gaussian BN of Example \ref{ex:BN} that the covariance matrix $\Sigma$ is perturbed by a $3\times 3$ matrix $D$ of all zeros except for a $d\in\mathbb{R}$ in positions $(2,1)$ and $(1,2)$ such that $\Sigma+D\in\R_{\text{\rm spsd}}^{3\times 3}$. The directed graph in Figure~\ref{fig:BN} is a faithful representation of this new Gaussian distribution if and only if the $2\times 2$ minor $(\sigma_{21}+d)\sigma_{32}-\sigma_{31}\sigma_{22}$ is equal to zero. But this is the case if and only if $d=0$: so if there is no perturbation. 

If alternatively the only non-zero entry of $D$ were in position $(1,1)$ then no matter what the value of $d\in\R$ the representation in Figure~\ref{fig:BN} would be a faithful description of the underlying conditional independence structure.
\end{example}

A possible approach to overcome the breaking of conditional independences in the case of Gaussian BNs is to vary the parameters of the univariate conditional Gaussian distributions in Definition~\ref{def:GBN}. The perturbation of the matrix $\Phi$ of conditional variances can then affect the covariance matrix of the overall Gaussian distribution \citep[see Section 6 of][for an example]{Gomez2013}. Another possibility is to perturb the matrix $B$ including the regression parameters and observe the effects of this perturbation on both the overall mean $\mu$ and covariance $\Sigma$. This second approach has been used to quantify the effect of adding or deleting edges in Gaussian BNs \citep{Gomez2011}. However and critically, both these approaches lose the intuitiveness of acting directly on the unconditional mean and covariance of the Gaussian distribution.

\subsection{Model-Preserving Sensitivity Analysis}
To overcome the difficulties arising in classical sensitivity analyses, we now introduce a novel approach which extends sensitivity methods usually applied exclusively to Gaussian BNs to more general Gaussian conditional independence models, including undirected Gaussian graphical models. In particular, we establish specific conditions under which a perturbed covariance matrix is within the original algebraic parameter set of the model at hand, so that all conditional independence relationships of the original model continue to be valid. We show below that this can be easily achieved by considering covariation schemes which act multiplicatively rather than additively on these  matrices. 

Henceforth, we  think of a Gaussian model $\M_{\CI}$ for a random vector $Y=(Y_i)_{i\in[n]}$ together with conditional independence assumptions $\CI=\{A_k\independent B_k~|~C_k \text{ for } k\in [r]\}$ as being represented by a collection of vanishing minors of its covariance matrix $\Sigma\in\PD$ as introduced in Lemma~\ref{prop:drton}. Because this rationale concerns only the covariance matrix, we can assume without loss  that $Y$ has zero mean, $\mu=0_n$. For ease of notation, we thus write $f_{\Sigma}$ rather than $f_{0_n,\Sigma}$ for its associated density.

We denote by the circle $\circ$ the \emph{Schur product} of two matrices $\Delta$ and $\Sigma$ of the same dimension, so $\Delta\circ\Sigma=(\delta_{ij}\sigma_{ij})_{i,j\in[n]}$ is the componentwise product of their entries. Let
\begin{equation}\label{eq:adjustment}
\Phi_{\Delta}: \Sigma\mapsto \Delta\circ\Sigma
\end{equation}
denote the map which sends a covariance matrix to its Schur product with a matrix $\Delta$. We  call the map $\Phi_{\Delta}$ \emph{model-preserving} if under this operation the algebraic parameter set in equation~(\ref{eq:parmset}) is mapped onto itself, $\Phi_{\Delta}(\A_{\CI})\subseteq\A_{\CI}$.

In the following sections, we always decompose the perturbation of a covariance matrix $\Sigma$ into two steps, and hence two Schur products, as follows. In the first step, the original covariance $\Sigma$ is mapped to its Schur product with a symmetric \emph{variation} matrix ${\Delta\in\mathbb{R}^{n\times n}_{\neq 0}}$. Hereby, usually only some of the original covariances $\sigma_{ij}$ are assigned a new value ${\sigma_{ij}\mapsto \delta_{ij}\sigma_{ij}}$ at selected positions $(i,j)$ while the remaining parameters are untouched. This is achieved by having all non-$(i,j)$ entries of $\Delta$ equal to one.  In demanding that all entries $\delta_{ij}$ are non-zero,
we automatically avoid setting a non-zero covariance $\sigma_{ij}\neq 0$ to zero via multiplication by an entry of $\Delta$. This type of perturbation would force the corresponding variables to be independent, $X_i\independent X_j$, in the perturbed model, which would clearly violate the assumptions in the original model $\M_{\CI}$.
The second Schur product is then calculated between $\Delta\circ\Sigma$ and a symmetric \emph{covariation} matrix $\tilde\Delta\in\mathbb{R}^{n\times n}_{\neq 0}$. This matrix $\tilde\Delta$ has ones in the positions $(i,j)$ whilst the values of the remaining entries need to be set according to some agreed procedure which ensures that for every vanishing minor of $\Sigma$, the appropriate minor of $\tilde\Delta\circ\Delta\circ\Sigma$ vanishes as well. In this process, in order to guarantee symmetry, whenever an entry $(i,j)$ is changed in one of the matrices, its corresponding entry $(j,i)$ needs to be changed in the exact same fashion. Explicitly, the composition of Schur products will be of the following form:
\begin{equation}\label{eq:decomposition}
\tilde{\Delta}\circ\Delta\circ\Sigma=
\begin{pmatrix}
\star&\cdots&\cdots&\star\\
\vdots&\ddots&1&\vdots\\
\vdots&1&\ddots&\vdots\\
\star&\cdots&\cdots&\star
\end{pmatrix}
\circ
\begin{pmatrix}
1&\cdots&\cdots&1\\
\vdots&\ddots&\delta_{ij}&\vdots\\
\vdots&\delta_{ji}&\ddots&\vdots\\
1&\cdots&\cdots&1
\end{pmatrix}
\circ
\begin{pmatrix}
\sigma_{11} &\cdots &\cdots & \sigma_{1n}\\
\vdots &\ddots&\sigma_{ij}&\vdots\\
\vdots&\sigma_{ji}&\ddots&\vdots\\
\sigma_{n1}&\cdots&\cdots&\sigma_{nn}
\end{pmatrix}
\end{equation}
Here, the stars indicate entries in $\tilde{\Delta}$ which need to be specified.
Thus, for a given covariance matrix $\Sigma\in\A_{\CI}$ and a given variation $\Delta\in\R_{\neq 0}^{n\times n}$ of that matrix, we now develop methods to find a covariation matrix $\tilde{\Delta}$ such that $\tilde{\Delta}\circ\Delta\circ\Sigma\in\A_{\CI}$. Then the map $\Phi_{\tilde{\Delta}\circ\Delta}$ is model-preserving. 

\begin{example}[Example \ref{ex:classicpert} continued]\label{ex:formal}
Suppose in the Gaussian model $Y_3\independent Y_2~|~Y_1$ we again perform a perturbation to the parameter $\sigma_{21}$ of the covariance matrix $\Sigma$. Then, under our formalism, the matrix $\Delta$ is defined as
\begin{equation}
\Delta=
\begin{pmatrix}
1&\delta&1\\
\delta&1&1\\
1&1&1
\end{pmatrix}
\end{equation}
and the only vanishing minor of $\Delta\circ\Sigma$ takes the form $\delta\sigma_{12}\sigma_{32}-\sigma_{31}\sigma_{22}$. This polynomial is equal to zero in either of three cases: when $\sigma_{22}$ is covaried by $\delta$; when $\sigma_{31}$ and $\sigma_{13}$ are covaried by $\delta$; or when $\sigma_{22}$, $\sigma_{31}$, $\sigma_{13}$, $\sigma_{32}$ and $\sigma_{23}$ are covaried by $\delta$. The associated covariation matrices $\tilde{\Delta}$ should equal, respectively,
\begin{equation}
\begin{pmatrix}
1&1&1\\
1&\delta&1\\
1&1&1
\end{pmatrix},
\qquad
\begin{pmatrix}
1&1&\delta\\
1&1&1\\
\delta&1&1
\end{pmatrix},
\qquad
\begin{pmatrix}
1&1&\delta\\
1&\delta&\delta\\
\delta&\delta&1
\end{pmatrix}.
\end{equation}
For each of these choices of $\tilde\Delta$, we have that $\Phi_{\tilde\Delta\circ\Delta}$ is model-preserving.

The structure associated to these perturbations is much clearer if we only consider the submatrix $\tilde\Delta_{\{2,3\},\{1,2\}}\circ\Delta_{\{2,3\},\{1,2\}}$ whose determinant is the relevant minor in Lemma~\ref{prop:drton}. For the three cases above, this matrix is equal to, respectively,
\begin{equation}
\begin{pmatrix}
\delta&\delta\\
1&1
\end{pmatrix},
\qquad
\begin{pmatrix}
\delta&1\\
\delta&1
\end{pmatrix},
\qquad
\begin{pmatrix}
\delta&\delta\\
\delta&\delta
\end{pmatrix}.
\end{equation}
So here the perturbation is applied either to a full row, a full column or the full matrix. We demonstrate below that this feature is in general associated to model-preservation.
\end{example}

The formalism we set up in this section enables us to interpret a model-preserving map as a homomorphism between polynomial rings in the indeterminates given by entries of the covariance and variation/covariation matrices. This observation together with Lemma~\ref{prop:drton} enables us to employ the powerful language of real algebraic geometry to study Gaussian conditional independence models. Over the next few sections, we make a first important step in using these notions for sensitivity analyses.

\section{One-Way Model-Preserving Sensitivity Analysis}
\label{sub:singlesa}

Throughout this section, we study single-parameter variations. Here, a user identifies precisely one entry $\sigma_{ij}$ of a covariance matrix at a fixed position $(i,j)$ that she intends to adjust to $\delta\cdot\sigma_{ij}$ for some $\delta\not=0,1$. The corresponding variation matrix $\Delta=(\delta_{ij})_{i,j}\in\R_{\neq0}^{n\times n}$ is a symmetric matrix which has at most two entries $\delta_{ij},\delta_{ji}=\delta$ not equal to one and entries $\delta_{kl}=1$ for all $(k,l)\not=(i,j)$.  We assume the user believes the conditional independence relationships of the model to be valid and that these should remain valid after the perturbation. 
Before setting up an appropriate covariation scheme for this setting, we fix some notation. This will be used in the two cases of our analysis presented below: first for models defined by a single conditional independence relationship and then for models defined by a collection of conditional independence statements.

For any symmetric matrix $D\in\R^{n\times n}$ and two index sets $A,B\subseteq[n]$, we henceforth denote with $\lfloor D_{A,B}\rfloor^1$ the symmetric, full dimension $n\times n$ matrix where:
\begin{itemize}
\item  all positions indexed by $A$ and $B$ are equal to the corresponding entries in $D$;
\item entries not indexed by $A$ and $B$ are set to ensure symmetry;
\item all other entries are equal to one.
\end{itemize}
We also let $\mathbbm{1}_{A,B}$ be the matrix with all entries equal to one and with rows indexed by $A$ and columns indexed by $B$. 

\begin{example}
Let $D\in\mathbb{R}^{3\times 3}$ and suppose 
\[
D_{\{1,2\},\{2,3\}}=\begin{pmatrix}
1&\delta\\
1&\delta
\end{pmatrix}.
\]
Then 
\[
\left\lfloor D_{\{1,2\},\{2,3\}}\right\rfloor^{1}=
\begin{pmatrix}
1&1&\delta\\
1&1&\delta\\
\delta&\delta&1
\end{pmatrix}.
\]
\end{example}

We now define different types of covariation matrices which are motivated by those studied in Example~\ref{ex:formal}.

\begin{definition}
\label{def:totpartrow}
For a single-parameter variation matrix $\Delta$ with $\delta_{ij}=\delta_{ji}=\delta$, we say that the covariation matrix $\tilde\Delta$ is 
\begin{itemize}
\item \emph{total} if $\tilde\Delta\circ\Delta=\delta\mathbbm{1}_{[n],[n]}$;
\item \emph{partial} if $\tilde\Delta\circ\Delta=\lfloor\delta\mathbbm{1}_{A\cup C,B\cup C}\rfloor^1$.
\item \emph{row-based} if $\tilde\Delta\circ\Delta=\lfloor\delta\mathbbm{1}_{E,B\cup C}\rfloor^1$ for a subset $E\subseteq A\cup C$;
\item \emph{column-based} if $\tilde\Delta\circ\Delta=\lfloor\delta\mathbbm{1}_{A\cup C,F}\rfloor^1$ for a subset $F\subseteq B\cup C$.
\end{itemize}
\end{definition}

In words, the Schur product of a variation with a total covariation matrix is a matrix filled with $\delta$, and the Schur product of a variation with a partial covariation matrix is a matrix which only has a symmetric subblock filled with $\delta$ and entries equal to one otherwise.  Row-based and column-based covariation matrices result in Schur products which have $\delta$ entries only in some specific subsets of the rows and columns. An illustration of such matrices were given in Example \ref{ex:formal}

By construction total, partial, row- and column-based covariations ensure symmetry. Henceforth, we assume that the perturbed matrix $\tilde\Delta\circ\Delta\circ\Sigma$ is also positive semidefinite, so that $\tilde\Delta\circ\Delta\circ\Sigma\in\PD$. We defer to the discussion at  the end of this paper for further details  on this assumption.

\subsection{One Conditional Independence Statement}
\label{sec:oneCI}
We first consider the case where a Gaussian conditional independence model $\M_{\CI}$ is specified by a single relationship $\CI=\{A\independent B~|~C\}$ for some index sets $A,B,C\subset[n]$. 
Throughout, $\Sigma\in\A_{\CI}$ is a covariance matrix in this model and $\Delta$ is a single-parameter variation matrix with non-one entry $\delta_{ij}=\delta_{ji}=\delta$ at a fixed position $(i,j)$ and $(j,i)$. We can now specify covariation matrices for this setup which result in model-preserving perturbations.

\begin{proposition}
\label{prop:1}
If $(i,j),(j,i)\not\in (A\cup C,B\cup C)$ then the map $\Phi_{\tilde\Delta\circ\Delta}$ is model-preserving for a covariation $\tilde\Delta = \mathbbm{1}_{[n],[n]}$.
\end{proposition}

This result is a straightforward consequence of Lemma~\ref{prop:drton}. Indeed, if both $(i,j)$ and $(j,i)$ are not entries of the submatrix $\Sigma_{A\cup C,B\cup C}$ whose vanishing minors specify the model, then no changes induced by multiplication with $\delta$ appear in the vanishing polynomials. This was illustrated in Example~\ref{ex:classicpert} for the variation of the variance $\sigma_{11}$. Proposition~\ref{prop:1} thus formalizes the cases when a perturbation has no effect on the underlying conditional independence structure.

\begin{proposition}
\label{prop:2}
If $C=\emptyset$ then the map $\Phi_{\tilde\Delta\circ\Delta}$ is model-preserving for $\tilde\Delta = \mathbbm{1}_{[n],[n]}$.
\end{proposition}

This result easily follows by noting that standard independence statements $A\independent B$ correspond to zeros in the covariance matrix. Multiplication of such zeros by $\delta$ still returns zeros which automatically results in a model-preserving map. Thus, if the model consists of one standard independence statement, any perturbation is model-preserving.  For standard sensitivity methods which act additively on the covariance matrix, such a property does not in general hold.

We next focus on the case where  a perturbation makes some of the original vanishing polynomials non-equal to zero. Henceforth, unless otherwise stated, we thus assume that either $(i,j)$ or $(j,i)$ are an entry of $\Sigma_{A\cup C,B\cup C}$. 

\begin{theorem}
\label{theo:total}
The map $\Phi_{\tilde\Delta\circ\Delta}$ is model-preserving for total and partial covariation matrices~$\tilde\Delta$.
\end{theorem}

A proof of this result is given in the appendix.

Observe that by default, we need to enforce that $\delta>0$ for total covariation matrices. This is because otherwise  the entries of the diagonal of $\Sigma$, the variances of the model, become negative. For partial covariations this constraint may not have to be enforced, even though in general it is very rare to investigate the effect of changing the sign of an entry in a covariance matrix. Furthermore,  there has been a growing interest on covariance matrices with the property that all their entries are positive \citep{Fallat2017,Slawski2015}.

As a consequence of Proposition~\ref{prop:1}, perturbations by $\delta$ outside of the submatrix $\Sigma_{A\cup C,B\cup C}$ in total covariation matrices have no effect on the vanishing polynomials. Henceforth we thus consider only the submatrix $\tilde\Delta_{A\cup C,B\cup C}$ and identify the entries that need to have a $\delta$ so that the map $\Phi_{\tilde\Delta\circ\Delta}$ is model-preserving.  Intuitively, this approach fits a user who may want to change the least possible number of entries of a covariance matrix after a perturbation. In Section~\ref{sect:KL} we give a theoretical justification for this.

\begin{theorem}
\label{theo:1}
The map $\Phi_{\tilde\Delta\circ\Delta}$ is model-preserving in the following cases:
\begin{itemize}
\item if $(i,j)$ or $(j,i) \in (A,B)$ for a row-based covariation $\tilde\Delta$ whenever $i\in E\subseteq A$, and for a column-based covariation $\tilde\Delta$ whenever $j\in F\subseteq B$;
\item if $(i,j)$ or $(j,i) \in (A,C)$ for a row-based covariation $\tilde\Delta$ whenever $i\in E\subseteq A$, and for a column-based covariation $\tilde\Delta$ whenever $F=C$;
\item if $(i,j)$ or $(j,i) \in (C,B)$ for a row-based covariation $\tilde\Delta$ whenever $E=C$, and for a column-based covariation $\tilde\Delta$ whenever $i\in F\subseteq B$;
\item if $(i,j)$ and $(j,i) \in (C,C)$ for a row-based covariation $\tilde\Delta$ whenever $E=C$, and for a column-based covariation $\tilde\Delta$ whenever $F=C$.
\end{itemize}
\end{theorem}

A proof of this result is given in the appendix.

In words, whenever the perturbed entry $(i,j)$ or $(j,i)$ is not an element of the conditioning set $(C,C)$, a row- or a column-based covariation consisting of one row or column only can give a model-preserving map $\Phi_{\tilde\Delta\circ\Delta}$. Conversely, if the entry $(i,j)\in(C,C)$ is in the conditioning set then row and column-based covariation matrices have $\delta$ entries over all rows and columns in $C$, respectively. This is because if for instance one row of $\tilde\Delta_{C,C}$ has $\delta$ elements then other entries of $\tilde\Delta_{C,C}$ need to be equal to $\delta$  to ensure symmetry. However this then implies that other full rows of $\tilde\Delta_{C,C}$ need to have all $\delta$ entries. An illustration of this can be found in Example \ref{ex:8} below.

It is possible that a model-preserving map $\Phi_{\tilde\Delta\circ\Delta}$ makes minors of other submatrices of $\Sigma$, possibly describing a conditional independence statement, vanish. In this case it is still true that $\Phi_{\tilde\Delta\circ\Delta}(\mathcal{A}_{\CI})\subseteq \mathcal{A}_{\CI}$ and that the original underlying graphical representation of the conditional independence model, if present, is still valid. However, this representation might not be minimal in the sense that it contains the smallest possible number of edges. Conversely, if $\Phi_{\tilde\Delta\circ\Delta}$ is not model-preserving, then the original underlying graphical representation is not valid.

\subsection{Multiple Conditional Independence Statements}
We now generalize the results of the previous section by considering models which are defined by a collection of multiple conditional independence relationships. Using the notation introduced in Section~\ref{sec:background}, in the following let $\CI=\{{A_1\independent B_1~|~C_1}, \ldots, {A_r\independent B_r~|~C_r}\}$.  Let also always $A=\cup_{k\in [r]}A_k$, $B=\cup_{k\in[r]}B_k$ and $C=\cup_{k\in[r]}C_k$.

First we introduce a result which simplifies the task of checking whether a covariation is model-preserving or not. Suppose hereby without loss of generality that the conditional independence relationships defining the model are ordered such that for all for $k\in [t]$ we have proper conditional independence statements $A_k\independent B_k~|~C_k$ where $C_k\neq\emptyset$, whilst for all $l\in[r]\setminus [t]$ the conditioning set is empty, $C_l=\emptyset$,  for some index for $t\leq r$. We then denote by $\CI^*=\{A_1\independent B_1~|~C_1,\dots,A_t\independent B_t~|~C_t\}\subseteq \CI$ the set of statements with non-empty conditioning set.

\begin{proposition}
If the map $\Phi_{\tilde\Delta\circ\Delta}$ is model-preserving for $\M_{\CI^*}$ then it is also model-preserving for $\M_{\CI}$.
\end{proposition}

This result is a straightforward consequence of Proposition~\ref{prop:2}, since zero entries in the covariance matrix are not affected by our model-preserving covariation. It is extremely useful since we can check whether a map is model-preserving by using only a subset of all conditional independences of a Gaussian model. Henceforth, we can thus without loss assume that the set $\CI$ is such that $C_i\neq\emptyset$ for all $i\in[r]$. 

The following example shows that in general it is not simply sufficient to create a $\tilde\Delta$ matrix for each conditional independence statement independently.

\begin{example}
\label{example:7}
Consider a model for $Y=(Y_i)_{i\in[5]}$ defined by 
$Y_4\independent Y_{\{1,2\}}~|~Y_3$ and $Y_{\{2,4\}}\independent Y_5| Y_3$. The submatrices associated to these independence statements are, respectively,
\begin{equation}
\begin{pmatrix}
\sigma_{31}&\sigma_{32}&\sigma_{33}\\
\sigma_{41}&\sigma_{42}&\sigma_{43}
\end{pmatrix}
\quad \text{and} \quad
\begin{pmatrix}
\sigma_{23}& \sigma_{25}\\
\sigma_{33}&\sigma_{35}\\
\sigma_{43}&\sigma_{45}
\end{pmatrix}
.
\end{equation}
Suppose the entry $\sigma_{43}$ of $\Sigma$ is varied by $\delta$ and that for both conditional independences the matrices $\tilde\Delta$ are column-based and consisting of one column only. If we compute the $\circ$ product between $\Delta$ and the two column-based $\tilde\Delta$ we have
\begin{equation}
\begin{pmatrix}
1&1&1&1&1\\
1&1&1&1&1\\
1&1&\delta&1&1\\
1&1&1&1&1\\
1&1&1&1&1
\end{pmatrix}
\circ
\begin{pmatrix}
1&1&1&1&1\\
1&1&\delta&1&1\\
1&\delta&\delta&1&1\\
1&1&1&1&1\\
1&1&1&1&1
\end{pmatrix}
\circ
\begin{pmatrix}
1&1&1&1&1\\
1&1&1&1&1\\
1&1&1&\delta&1\\
1&1&\delta&1&1\\
1&1&1&1&1
\end{pmatrix}
=
\begin{pmatrix}
1&1&1&1&1\\
1&1&\delta&1&1\\
1&\delta&\delta^2&\delta&1\\
1&1&\delta&1&1\\
1&1&1&1&1
\end{pmatrix}
\end{equation}
When the above matrix is then multiplied with $\Sigma$ we have that, for instance, the minor $\delta\sigma_{31}\sigma_{43}-\delta^2\sigma_{41}\sigma_{33}\neq 0$ does not vanish and thus the resulting map is not model-preserving.
\end{example}

The problem here is that the entry $\sigma_{33}$ appears in both submatrices whose minors need to vanish. This observation leads us to the following definition and subsequent result.

\begin{definition}
We say that two relationships $A_k\independent B_k~|~ C_k$ and $A_l\independent B_l~|~C_l$ in a Gaussian conditional independence model $\M_{\CI}$ are \emph{separated} if for any entry $\sigma_{kl}$ of $\Sigma_{A_k\cup C_k,B_k\cup C_k}$ neither $\sigma_{kl}$ nor $\sigma_{lk}$ are in $\Sigma_{A_l\cup C_l,B_l\cup C_l}$ and viceversa. A model $\M_{\CI}$ is called \emph{separable} if all its pairs of conditional independence statements are separated.
\end{definition}

Intuitively, separated statements will be described by collections of vanishing minors which can be specified independently.

\begin{proposition}
Let $\M_{\CI}$ be separable with parameter space $\A_{\CI}$.
Given a covariance matrix $\Sigma\in\A_{\CI}$, the map $\Phi_{\tilde\Delta\circ\Delta}$ is model-preserving for $\tilde{\Delta}=\circ_{k\in[r]}\tilde{\Delta}_k$, where $\tilde{\Delta}_k$ is a $\M_{\CI_k}$-preserving covariation matrix for $\CI_k=\{A_k\independent B_k~|~C_k\}$.
\end{proposition}

The result easily follows by separability of $\M_{\CI}$.

We can now consider generic, non-separable, Gaussian conditional independence models and study covariation matrices associated to their model-preserving maps. In direct analogy to Proposition~\ref{prop:1} and Theorem~\ref{theo:total}, we find the following.

\begin{proposition}
The map $\Phi_{\tilde\Delta\circ\Delta}$ is model-preserving for
\begin{itemize}
\item $\tilde{\Delta}=\mathbbm{1}_{[n],[n]}$ if $(i,j),(j,i)\not\in (A\cup C,B\cup C)$;
\item $\tilde{\Delta}=\mathbbm{1}_{[n],[n]}$ if $(i,j),(j,i)\not\in (A_k\cup C_k,B_k\cup C_k)$ for all $k\in[r]$;
\item total and partial covariation matrices $\tilde{\Delta}$.
\end{itemize}
\end{proposition}

The first two statements easily follow by noting that no changes appear in any of the vanishing polynomials. The third point is a straightforward consequence of Theorem~\ref{theo:total}.

Next we again look for covariation matrices which include a smaller number of elements than total and partial covariation matrices. This generalizes the concept of row-based and column-based covariations from models defined by single conditional independences to models defined by multiple relationships. Following the results of Section~\ref{sec:oneCI}, it is reasonable to consider simplifications of partial covariation matrices where some of the rows/columns have entries equal to one. Thus we study covariation for the submatrix $\Sigma_{A\cup C,B\cup C}$. The following example illustrates some of the difficulties we might encounter.

\begin{example}[Example \ref{example:7} continued]
\label{ex:8}
For the Gaussian model defined by $Y_4\independent Y_{\{1,2\}}~|~Y_3$ and $Y_{\{2,4\}}\independent Y_5~|~Y_3$ where we varied the covariance $\sigma_{43}$, we need to consider the submatrix $\Sigma_{\{2,3,4\},\{1,2,3,5\}}$. Simple row-based and  column-based covariations are associated to the matrices  $\tilde\Delta_{\{2,3,4\},\{1,2,3,5\}}\circ\Delta_{\{2,3,4\},\{1,2,3,5\}}$ corresponding to
\begin{equation}
\label{eq:example}
\begin{pmatrix}
1&1&1&1\\
1&1&1&1\\
\delta&\delta&\delta&\delta
\end{pmatrix}
\quad\text{and}\quad
\begin{pmatrix}
1&1&\delta&1\\
1&1&\delta&1\\
1&1&\delta&1
\end{pmatrix},
\end{equation}
respectively.

For the row-based covariation on the left of equation (\ref{eq:example}) we have 
\[
\tilde\Delta_{\{2,3,4\},\{1,2,3,5\}}\circ\Delta_{\{2,3,4\},\{1,2,3,5\}}=(\lfloor\tilde\Delta_{\{2,3,4\},\{1,2,3,5\}}\circ\Delta_{\{2,3,4\},\{1,2,3,5\}}\rfloor^1)_{\{2,3,4\},\{1,2,3,5\}}
\]
 since $\sigma_{14}$, $\sigma_{24}$, $\sigma_{34}$ and $\sigma_{54}$ are not in $\Sigma_{\{2,3,4\},\{1,2,3,5\}}$. This means that no entries of $(\tilde\Delta\circ\Delta)_{\{2,3,4\},\{1,2,3,5\}}$ are altered during the creation of the $5\times 5$ matrix $\lfloor\tilde\Delta\circ\Delta\rfloor^1$. It is straightforward to check that this covariation matrix gives rise to a model-preserving map. Conversely, consider the  column-based covariation on the right of equation~(\ref{eq:example}). 
In this case, $\tilde\Delta_{\{2,3,4\},\{1,2,3,5\}}\circ\Delta_{\{2,3,4\},\{1,2,3,5\}}\neq(\lfloor\tilde\Delta_{\{2,3,4\},\{1,2,3,5\}}\circ\Delta_{\{2,3,4\},\{1,2,3,5\}}\rfloor^1)_{\{2,3,4\},\{1,2,3,5\}}$ because
\begin{equation}
(\lfloor\tilde\Delta_{\{2,3,4\},\{1,2,3,5\}}\circ\Delta_{\{2,3,4\},\{1,2,3,5\}}\rfloor^1)_{\{2,3,4\},\{1,2,3,5\}}=\begin{pmatrix}
1&1&\delta&1\\
1&\delta&\delta&1\\
1&1&\delta&1
\end{pmatrix}.
\end{equation} 
The map based on such a covariation is not model-preserving. This is because covariation matrices need to be filled with full-row or full-columns of $\delta$s in order to preserve a model's structure, as demonstrated in Theorem~\ref{theo:1}. We can fix this issue by simply filling up the second column of that matrix with $\delta$ entries. Indeed,
\begin{equation}
\tilde\Delta_{\{2,3,4\},\{1,2,3,5\}}\circ\Delta_{\{2,3,4\},\{1,2,3,5\}}	=\begin{pmatrix}
1&\delta&\delta&1\\
1&\delta&\delta&1\\
1&\delta&\delta&1
\end{pmatrix}.
\end{equation}
gives a model-preserving map because $\sigma_{24}$ is not an entry of $\Sigma_{\{2,3,4\},\{1,2,3,5\}}$.
\end{example}

The above example demonstrated that again row-based and column-based covariations can be associated to model-preserving maps. We can thus generalize Theorem~\ref{theo:1} to the following result.

\begin{proposition}
\label{prop:aereo}
The map $\Phi_{\tilde\Delta\circ\Delta}$ is model-preserving for a row-based or a column-based covariation matrix $\tilde{\Delta}$ if 
\begin{equation}
\label{eq:cond}
\tilde\Delta_{A\cup C,B\cup C}\circ\Delta_{A\cup C,B\cup C}=(\lfloor\tilde\Delta_{A\cup C,B\cup C}\circ\Delta_{A\cup C,B\cup C}\rfloor^1)_{A\cup C,B\cup C}.
\end{equation}
\end{proposition}

This result easily follows by noting that under the condition in equation~(\ref{eq:cond}) the map $\Phi_{\tilde{\Delta}\circ\Delta}$ is model-preserving for each $\M_{\CI^k}$ with $\CI=\{A_k\independent B_k~|~C_k\}$, since by construction every submatrix $(\tilde\Delta\circ\Delta)_{A_k\cup C_k,B_k\cup C_k}$ is a row-based or column-based covariation matrix. In other words, model-preservation is guaranteed if by creating the full-dimensional covariation matrix no entries with indexes in $A\cup C$ and $B\cup C$ are affected to ensure symmetry of the resulting matrix.

\section{Multi-Way Model-Preserving Sensitivity Analysis}
\label{sec:multi}

We can now generalize the results of Section~\ref{sub:singlesa} by studying multi- rather than single-parameter variations in Gaussian conditional independence models. In particular, we show below that the characterization of parameter sets as sets of vanishing polynomial equations provide a powerful language to straightforwardly tackle this much more general case.

\begin{theorem}\label{theo:compositions}
Compositions of model-preserving maps are model-preserving. In particular, for any two matrices $\Delta$ and $\Delta'$ we have $\Phi_\Delta(\Phi_{\Delta'})=\Phi_{\Delta\circ \Delta'}$.
\end{theorem}

A proof of this result is given in the appendix.

Theorem~\ref{theo:compositions} immediately implies that if $\tilde{\Delta}\circ\Delta$ is a model-preserving covariation scheme then any further model-preserving covariation $\tilde{\Delta}'\circ\Delta'\circ\tilde{\Delta}\circ\Delta$ does not violate the conditional independences of the model. This implies that parameters can be varied sequentially.

In fact, we can write any symmetric  multi-way variation matrix $\Delta$ as the Schur product of matrices of the form considered in  Section~\ref{sub:singlesa}, namely matrices $\Delta^k$ of ones with at most two entries $\delta_{ij}^k=\delta_{ji}^k$ different from one and not equal to zero. In this notation we use superscripts in order to avoid double indices. Explicitly, we then have $\Delta=\Delta^{1}\circ \Delta^{2} \circ \cdots \circ \Delta^{n}$ where every $\Delta^{k}$ enforces a single-parameter variation. We can now covary every single-parameter variation $\Delta^{k}$ by a matrix $\tilde{\Delta}^{k}$ using for instance row-based and column-based covariation matrices as in Proposition~\ref{prop:aereo}. Because the Schur product is commutative, this induces a map
\begin{equation}\label{eq:multivar}
\Phi_{\tilde{\Delta}^{1}\circ \Delta^{1}\circ\tilde{\Delta}^{2}\circ \Delta^{2} \circ \cdots \circ \tilde{\Delta}^{n}\circ\Delta^n}=\Phi_{\tilde{\Delta}^{1}\circ\tilde{\Delta}^{2}\circ\cdots \circ \tilde{\Delta}^{n}\circ \Delta^{1}\circ \Delta^{2}\circ\cdots\circ \Delta^{n}}=\Phi_{\tilde{\Delta}\circ\Delta}
\end{equation}
where $\tilde{\Delta}=\tilde{\Delta}^{1}\circ \tilde{\Delta}^{2} \circ \cdots \circ \tilde{\Delta}^{n}$ is the covariation matrix for $\Delta$.
By Theorem~\ref{theo:compositions}, this map is model-preserving.

\begin{example}[Example \ref{ex:8} continued]
For the Gaussian model defined by $Y_4\independent Y_{\{1,2\}}~|~Y_3$ and $Y_{\{2,4\}}\independent Y_5~|~Y_3$ suppose that not only the covariance $\sigma_{43}$ is varied by a quantity $\delta_1$, but also the entry $\sigma_{32}$ is varied by $\delta_2$. From Example \ref{example:7} we know that the row-based covariation matrix on the left hand side of equation (\ref{eq:example}) is model-preserving for the variation by $\delta_1$. From Proposition \ref{prop:aereo} it easily deduced that 
\begin{equation}
(\tilde{\Delta}^2\circ \Delta^2)_{\{2,3,4\},\{1,2,3,5\}}= \begin{pmatrix}
1&\delta_2&\delta_2&1\\
1&\delta_2&\delta_2&1\\
1&\delta_2&\delta_2&1
\end{pmatrix}
\end{equation}
is associated to a model-preserving covariation matrix. Therefore, using Theorem \ref{theo:compositions} we can construct the matrix
\begin{equation}
\tilde{\Delta}\circ\Delta=\begin{pmatrix}
1&1&1&\delta_1&1\\
1&\delta_2&\delta_2&\delta_1\delta_2&1\\
1&\delta_2&\delta_2&\delta_1\delta_2&1\\
\delta_1&\delta_1\delta_2&\delta_1\delta_2&1&\delta_1\\
1&1&1&\delta_1&1
\end{pmatrix}
\end{equation}
which is associated to a model-preserving map.
\end{example}

\section{Divergence Quantification}
\label{sect:KL}
The previous sections formalized how variations of the covariance matrix of a Gaussian model can be coherently performed without affecting its conditional independence structure. Next, as usual in sensitivity studies, we quantify the dissimilarity between the original and the new distribution. We start by considering the KL divergence.

Using notation from Section~\ref{sec:stan}, let $Y$ be a Gaussian vector with density $f_{\Sigma}$ and let $\tilde{Y}$ be the vector resulting from a model-preserving variation and having density $f_{\tilde{\Delta}\circ\Delta\circ\Sigma}$. Thus both covariance matrices belong to the parameter set of the same model, that is $\Sigma,\tilde{\Delta}\circ\Delta\circ\Sigma\in\A_{\CI}$.  Here, $\Delta$ and $\tilde{\Delta}$ may be associated to either single- or multi-parameter variations. 
In the latter case, as formalized in Section \ref{sec:multi}, we again denote $\Delta=\Delta^1\circ\cdots\circ\Delta^n$ as a Schur product of matrices $\Delta^k$ associated to a single-parameter variation and by $\tilde{\Delta}^k$ their model-preserving covariation matrix. Then $\tilde{\Delta}=\tilde{\Delta}^1\circ\cdots\circ\tilde{\Delta}^n$. Let $\delta_k$ be the variation associated to the matrix $\Delta^k$. The KL divergence between $Y$ and $\tilde{Y}$ in model-preserving sensitivity analyses can be written as

\begin{equation}
\label{eq:KLour}
\KL(\tilde{Y}||Y)=\dfrac{1}{2}\left[\tr(\Sigma^{-1}(\tilde{\Delta}\circ\Delta\circ\Sigma))-n+\log\dfrac{\det(\Sigma)}{\det(\tilde{\Delta}\circ\Delta\circ\Sigma)}\right].
\end{equation}

This result easily follows by substituting the definition of our variation and covariation matrices into equation~(\ref{eq:KL}). 

Whilst for partial and row/column-based covariations KL does not entertain a closed form, for total covariation matrices KL divergence has the following simple closed-form formula.
\begin{equation}
\label{equazione}
\KL(\tilde{Y}||Y)=\frac{1}{2}\bigl(n(\delta-\log(\delta)-1)\bigr)
\end{equation}
where $\delta=\prod_{i\in[n]}\delta_i$ for a multi-way variation. A derivation of this result can be found in the appendix.

Our examples in the next section demonstrate that KL divergences often behave counterintuitively and differently depending on the form of the covariance matrix analysed. Similar results were observed not only for KL divergences but also for other members of the class of $\phi$-divergences, for instance the inverse KL divergence and the Hellinger distance \citep{Ali1966}. For this reason we recommend using the KL divergence in conjunction with another measure which takes into account the number of entries that have been varied. One such measure is the Frobenius norm, defined below, which has been recently used in econometrics and finance to quantify the distance between two covariance matrices \citep{Amendola2015, Laurent2012}. 

\begin{definition}\label{def:frob}
Let $Y$ and $Y'$ be two Gaussian vectors with distribution $f_{\Sigma}$ and $f_{\Sigma'}$, respectively. The \emph{Frobenius norm} between $Y$ and $Y'$ is defined as
\begin{equation}
\F(Y,Y')=\tr((\Sigma-\Sigma')^{\top}(\Sigma-\Sigma')).
\end{equation}
\end{definition}

In words, the Frobenius norm is defined as the sum of the element-wise squared differences of the two covariance matrices. For standard sensitivity analyses where a variation matrix $D$ acts additively on $\Sigma$, the Frobenius norm is simply equal to $\tr(D^{\top}D)$, consisting of the sum of the squared variations. For our multiplicative covariation, we have the following result.

\begin{proposition}
\label{prop:frob}
Let $\tilde{\Delta}\circ\Delta=(\delta_{ij})_{ij}$ be model-preserving. Then
\begin{equation}
\F(Y,\tilde{Y})=\sum_{i,j\in[n]}(1-\delta_{ij})^2\sigma_{ij}^2.
\end{equation}
\end{proposition}

This result easily follows by substituting $\Sigma'=\tilde{\Delta}\circ\Delta\circ\Sigma$ into the formula given in Definition~\ref{def:frob}.

Proposition \ref{prop:frob} enables us to deduce a useful ranking based on the Frobenius norm of the various model-preserving covariation schemes we introduced. Letting $\tilde{Y}_{\text{total}}$, $\tilde{Y}_{\text{partial}}$, $\tilde{Y}_{\text{row}}$ and $\tilde{Y}_{\text{column}}$ be the random vectors resulting from total, partial, row-based and column-based covariations, respectively, the following inequalities hold:
\begin{equation}
\F(Y,\tilde{Y}_{\text{total}})\geq \F(Y,\tilde{Y}_{\text{partial}}), \quad \F(Y,\tilde{Y}_{\text{partial}})\geq \F(Y,\tilde{Y}_{\text{row}}), \quad \F(Y,\tilde{Y}_{\text{partial}})\geq \F(Y,\tilde{Y}_{\text{column}}).
\end{equation}
This is true simply because, by definition, total covariations affect more entries of the covariance matrix than partial ones.
Similarly, partial covariations affect more entries than row- and column-based covariations.

Since it is always possible to find a variation $d_{ij}$ that acts additively on $\sigma_{ij}$ such that $d_{ij}+\sigma_{ij}=\delta_{ij}\sigma_{ij}$, we can also deduce using the same reasoning that
\begin{equation}
\F(Y,\tilde{Y}_{\text{column}})\geq \F(Y,\tilde{Y}_{\text{standard}})\quad\text{and}\quad \F(Y,\tilde{Y}_{\text{row}})\geq \F(Y,\tilde{Y}_{\text{standard}}),
\end{equation}
where $\tilde{Y}_{\text{standard}}$ is the vector resulting from standard sensitivity methods which in general break the conditional independence structure of the model. Our examples in the following give an empirical illustration of the above inequalities.

\section{Illustrations}

 \begin{figure}
\centering
\begin{tikzpicture}
\renewcommand{\xx}{1.5}
\renewcommand{\yy}{0.8}
\node (x1) at (0,0) {1};
\node (x2) at (\xx,0) {2};
\node (x3) at (2*\xx,0) {3};
\node (x4) at (\xx,\xx){4};
\draw[->] (x1) -- (x2);
\draw[->] (x2) -- (x3);
\draw[->] (x1) -- (x4);
\draw[->] (x2) -- (x4);
\draw[->] (x3) -- (x4);
\end{tikzpicture}
\caption{The directed acyclic graph representing the conditional independence model in Section~\ref{sec:ex1}. \label{fig:BNex}}
\end{figure}
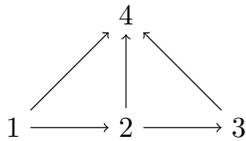

We now illustrate the results of the previous sections using two examples: one artificial and one based on a real-world data application.

\subsection{A First Example}
\label{sec:ex1}
Consider the BN model represented in Figure \ref{fig:BNex} and associated to the covariance matrix
\begin{equation}
\label{eq:cov}
\Sigma =\begin{pmatrix}
1&2&2&7\\
2&5&5&17\\
2&5&6&19\\
7&17&19&63
\end{pmatrix}.
\end{equation}
This matrix was deduced using the formalism of Section~\ref{sec:BN} by setting $\beta_{0i}=0$, $\sigma_i=1$, for $i\in[4]$, $\beta_{12}=2$, $\beta_{13}=0$, $\beta_{23}=1$, $\beta_{14}=1$, $\beta_{24}=1$ and $\beta_{34}=2$. Covariance matrices with a structure similar to the one in equation (\ref{eq:cov}) are often encountered when the $\beta_{ij}$ parameters are expert-elicited \citep[see for instance][]{Gomez2011,Gomez2013}. Notice that this BN is defined by only one conditional independence statement, namely $Y_3\independent Y_1~|~Y_2$. This is equivalent to the vanishing minor ${\sigma_{12}\sigma_{23}-\sigma_{22}\sigma_{13}=0}$ by Lemma~\ref{prop:drton}. Thus only variations of the parameters $\sigma_{21}$, $\sigma_{22}$, $\sigma_{31}$ and $\sigma_{32}$ may break the conditional independence structure of this model. 

\begin{figure}
\begin{center}
\includegraphics[scale=0.5]{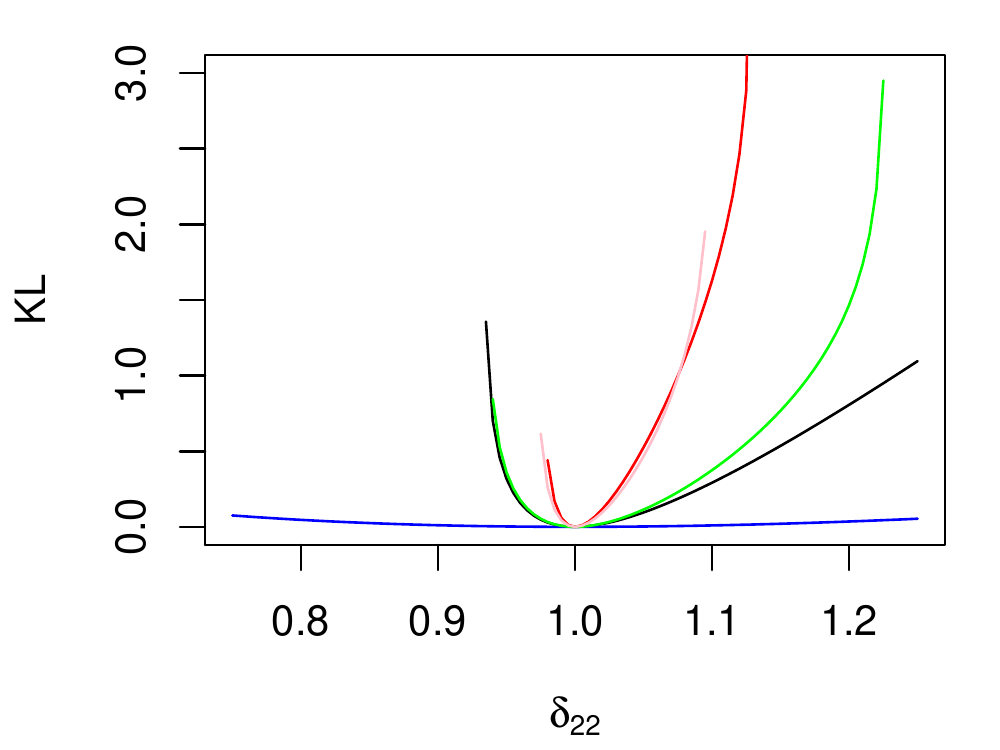}\;\;\;\;
\includegraphics[scale=0.5]{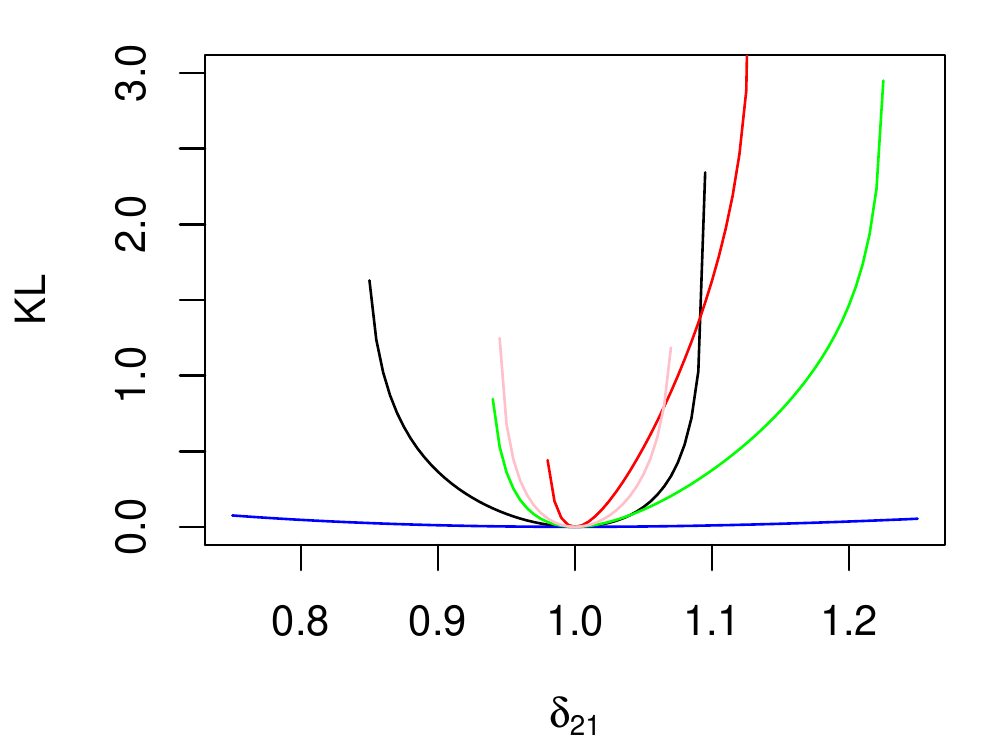}
\includegraphics[scale=0.5]{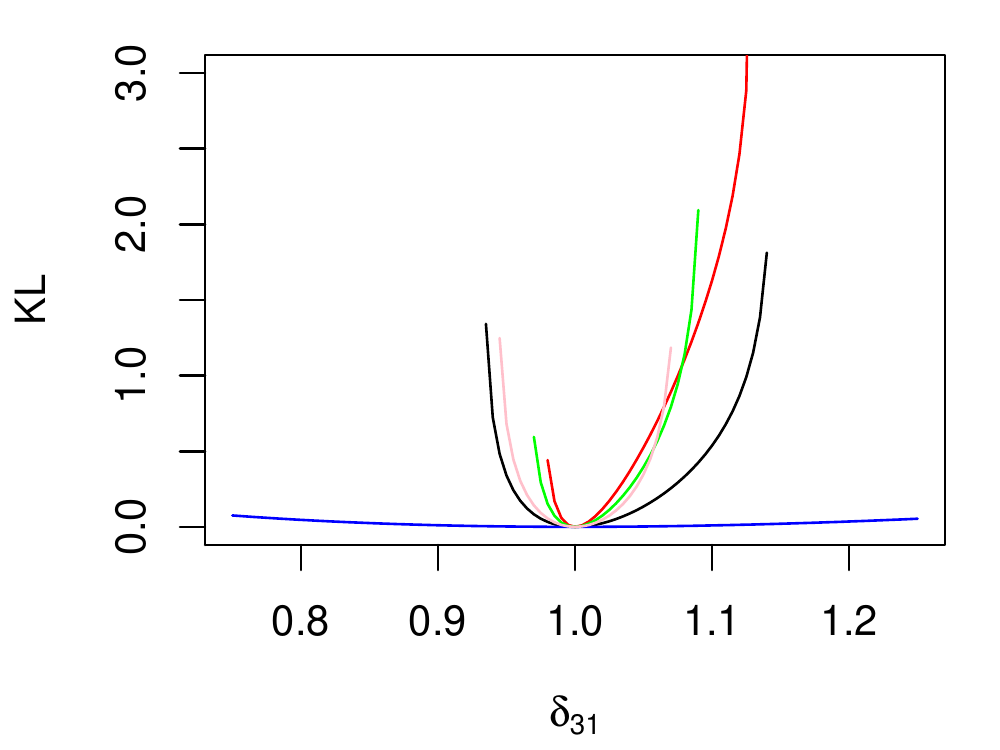}\;\;\;\;
\includegraphics[scale=0.5]{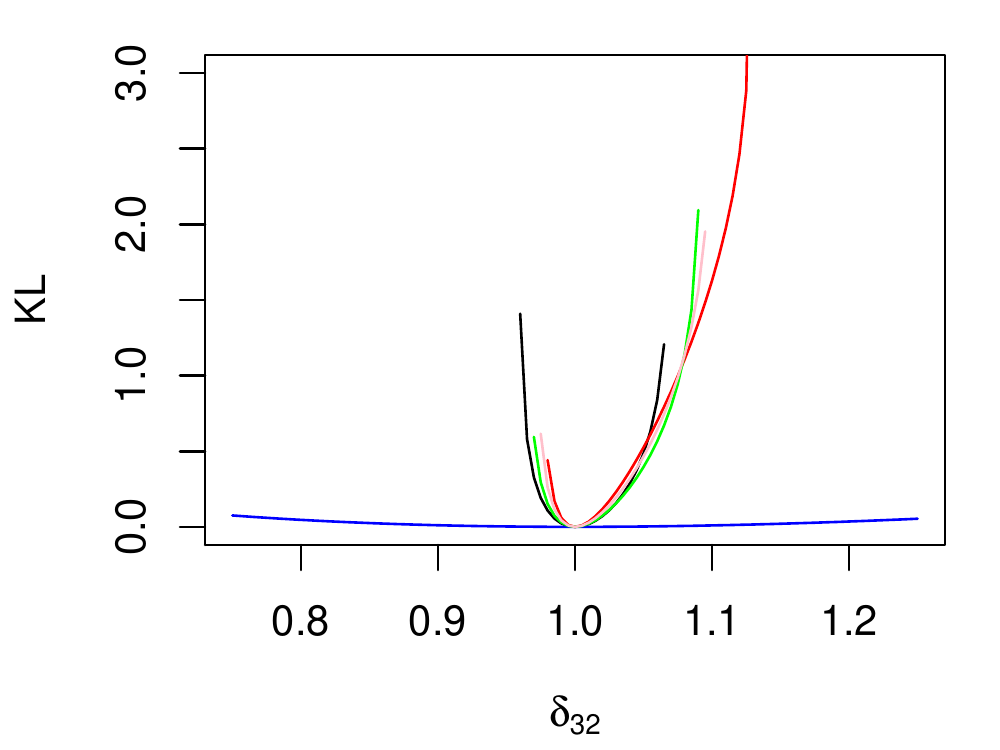}
\end{center}
\caption{KL divergence for one-way variations $\sigma_{ij}\mapsto\delta_{ij}\sigma_{ij}$ of the parameters of the network in Figure~\ref{fig:BNex}. We use the color codes black = standard variation; blue = full;  red = partial; green =  row-based; pink = column-based.\label{fig:KL}}
\end{figure}

Figure \ref{fig:KL} reports the KL divergence for one-way sensitivity analyses of each of the above parameters when entries are either increased or decreased by 25\%. The plots show that the KL divergence is considerably smaller for full-covariation matrices than for all the other covariations as well as for standard sensitivity methods. All other methods have similar KL divergences and we see that for most variations there is one model-preserving covariation with KL divergence either smaller or comparable to the one of the standard method. The KL divergence takes similar values for all parameters varied and therefore none of these has a predominant effect on the robustness of the network.

\begin{figure}
\begin{center}
\includegraphics[scale=0.5]{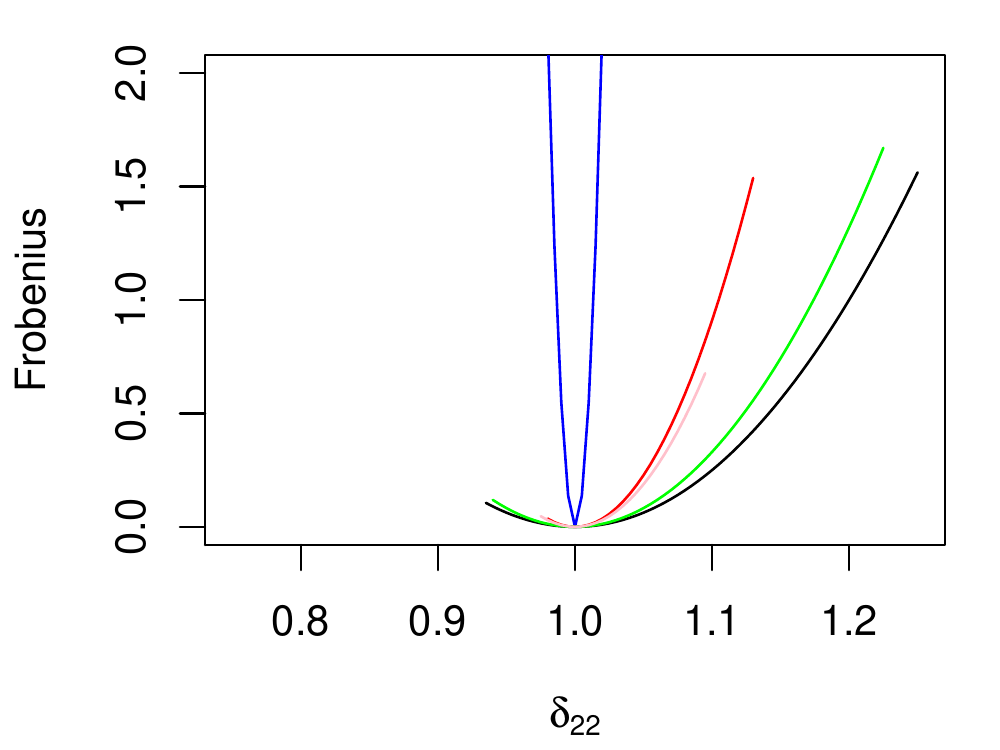}\;\;\;\;
\includegraphics[scale=0.5]{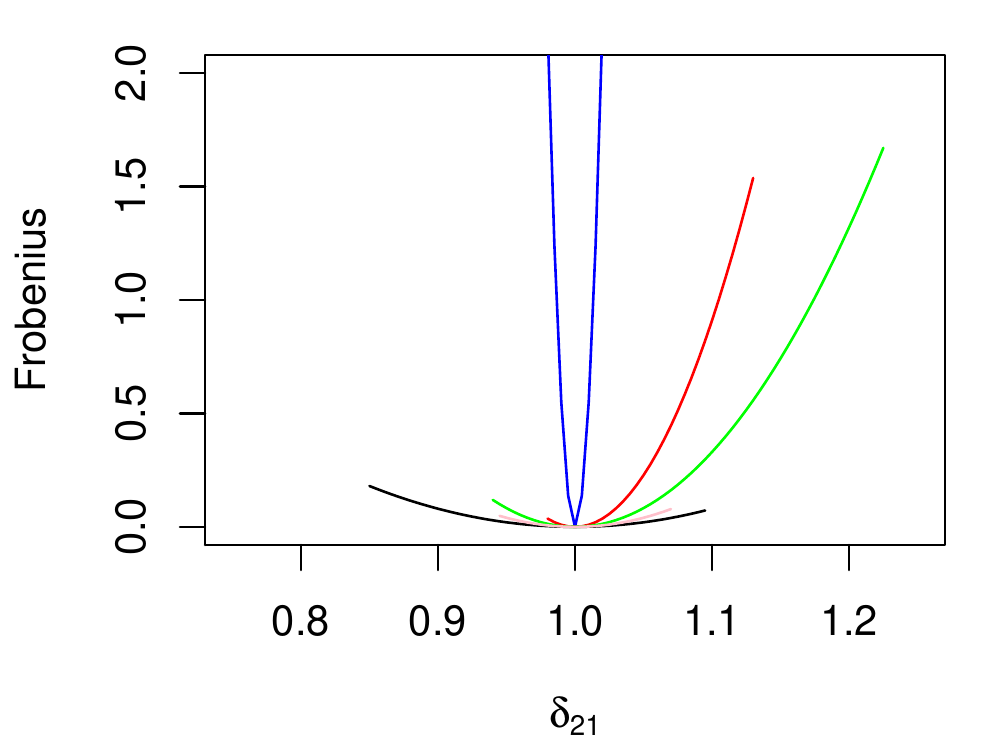}
\includegraphics[scale=0.5]{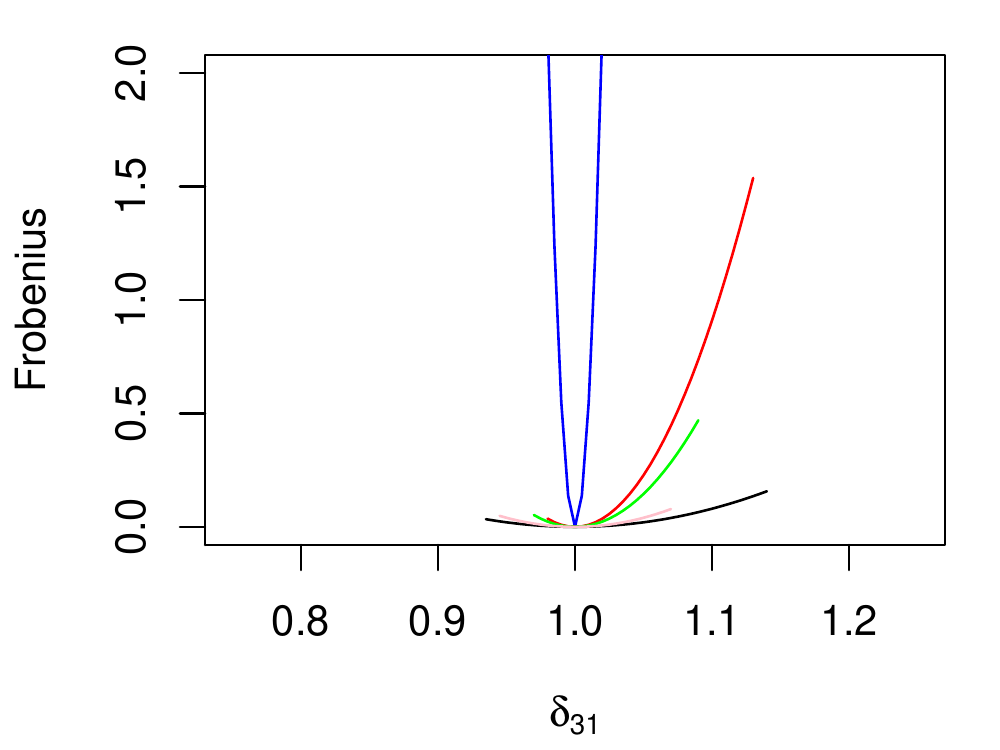}\;\;\;\;
\includegraphics[scale=0.5]{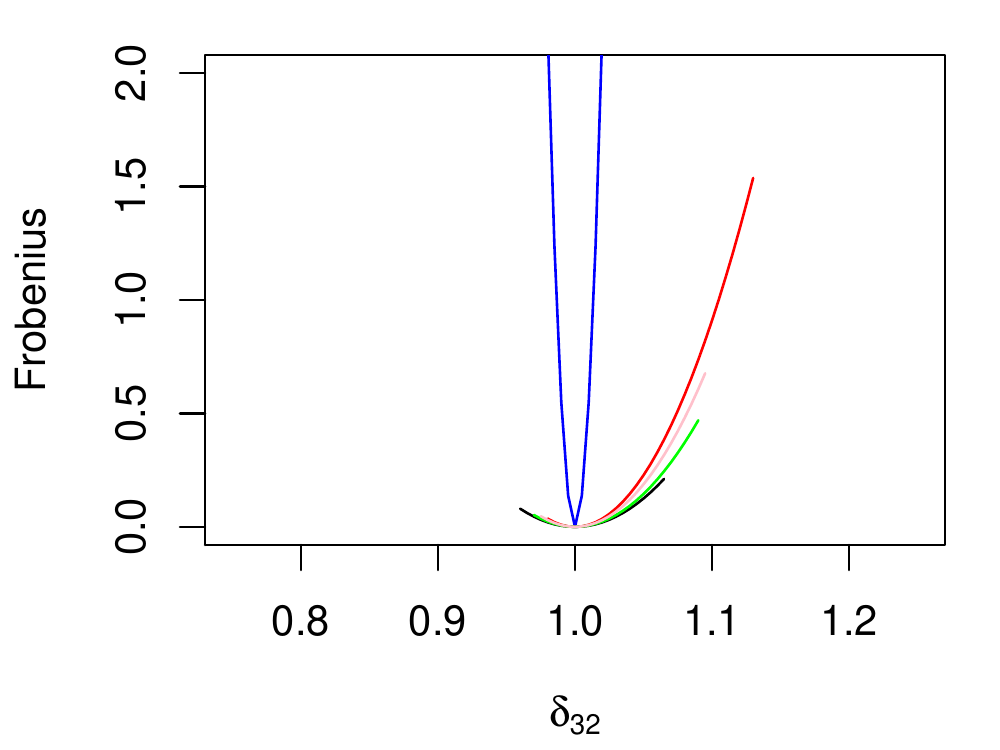}
\end{center}
\caption{Frobenius norm for one-way variations $\sigma_{ij}\mapsto\delta_{ij}\sigma_{ij}$ of the parameters of the network in Figure~\ref{fig:BNex}. We use the color codes black = standard; blue = full;  red = partial; green = row-based; pink = column-based.\label{fig:Fro}}
\end{figure}

Figure \ref{fig:Fro} reports the Frobenius norms under the same settings as above, confirming the theoretical results of Section \ref{sect:KL}. In particular, standard sensitivity methods always have a smaller Frobenius norm than the others  because in this case less parameters are varied. The plots also confirm that there is no fixed ranking between column-based and row-based covariations and demonstrate that  full covariation has a considerably larger Frobenius norm than the other approaches. Again, the Frobenius norm appears to be comparable between all parameters varied and therefore none of these seems to be critical.

\begin{figure}
\begin{center}
\includegraphics[scale=0.45]{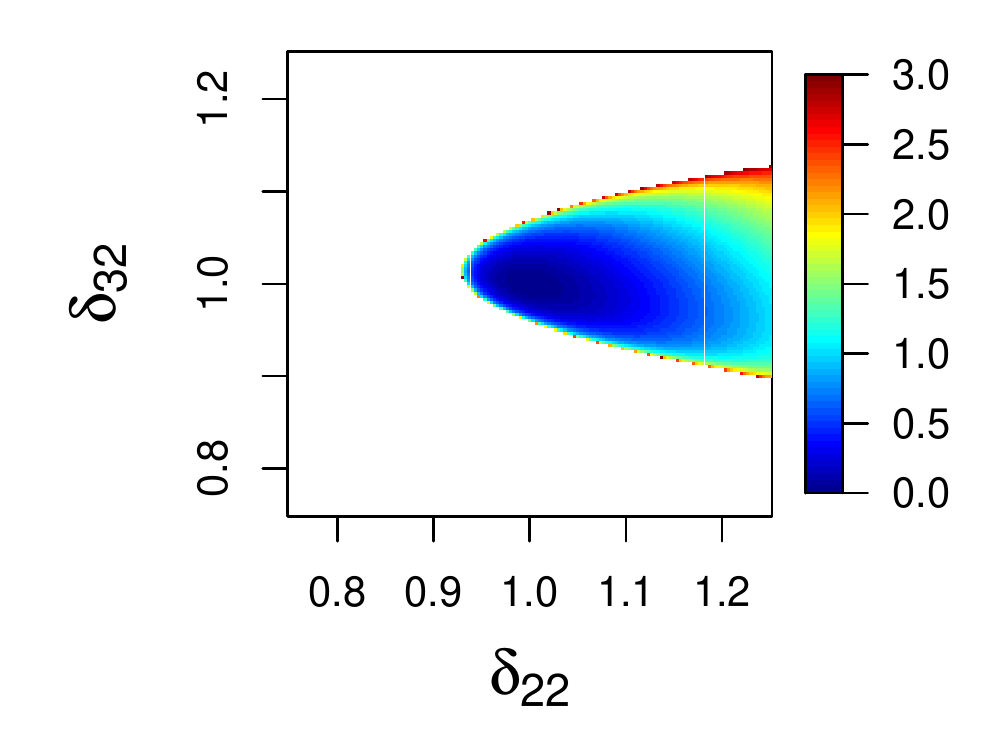}  \,\,\,\,\,\,\,\,
\includegraphics[scale=0.45]{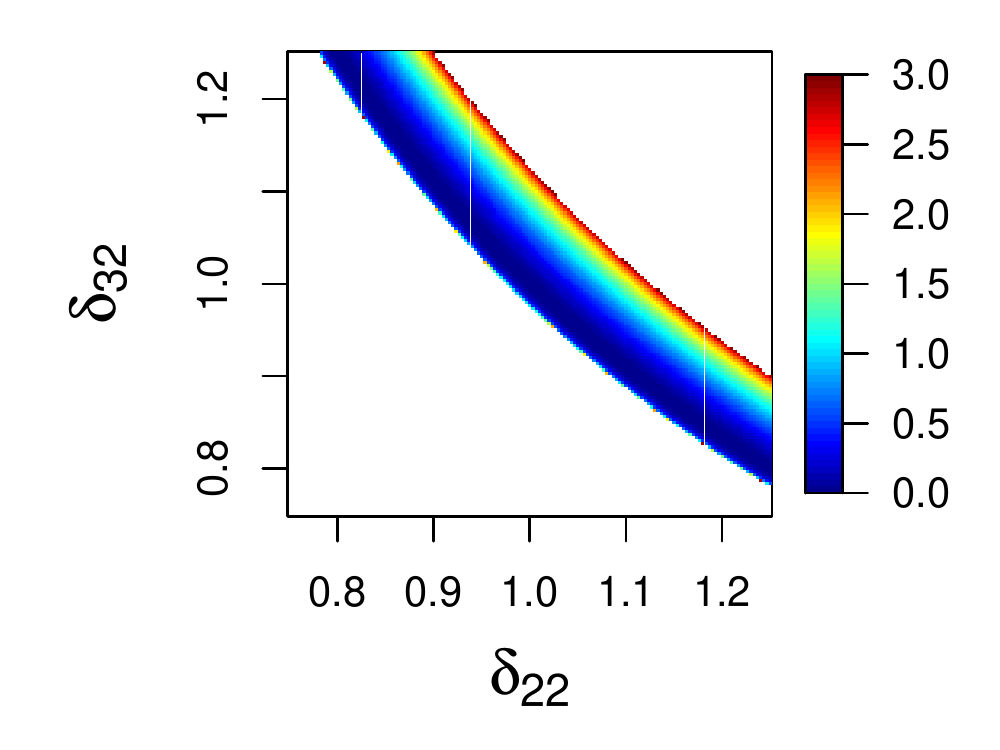}\,\,\,\,\,\,\,\,
\includegraphics[scale=0.45]{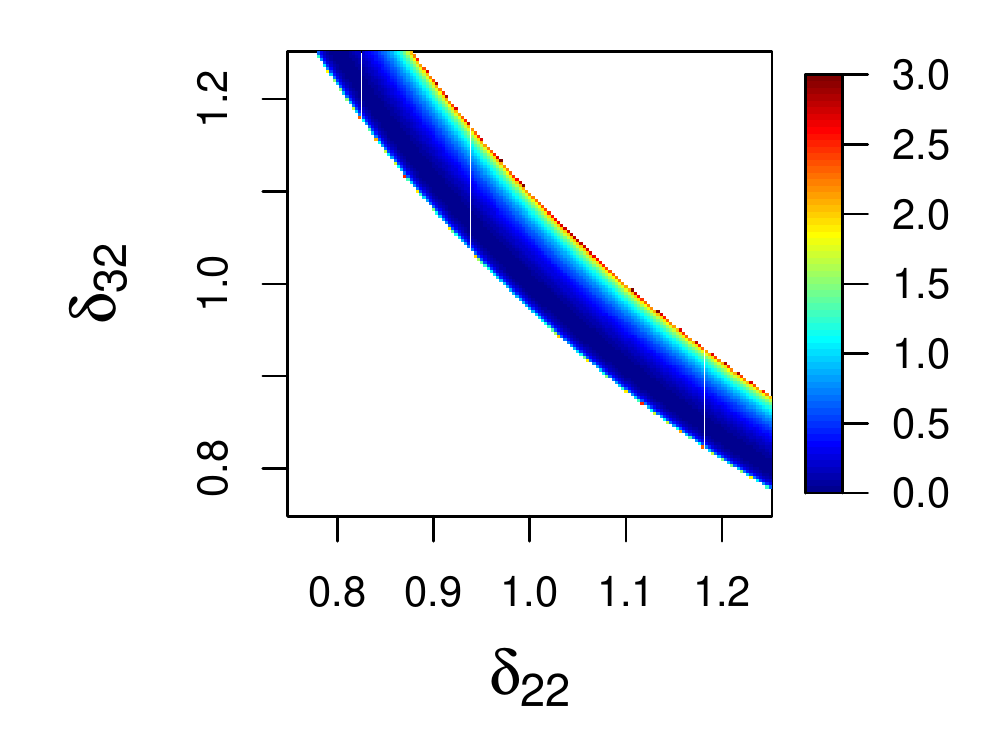}
\end{center}
\caption{KL divergence for multi-way variation of the parametes $\sigma_{22}$ and $\sigma_{32}$ of the parameters of the network in Figure \ref{fig:BNex}: standard (left); partial (central); column-based (right).\label{fig:KL2}}
\end{figure}

\begin{figure}
\begin{center}
\includegraphics[scale=0.45]{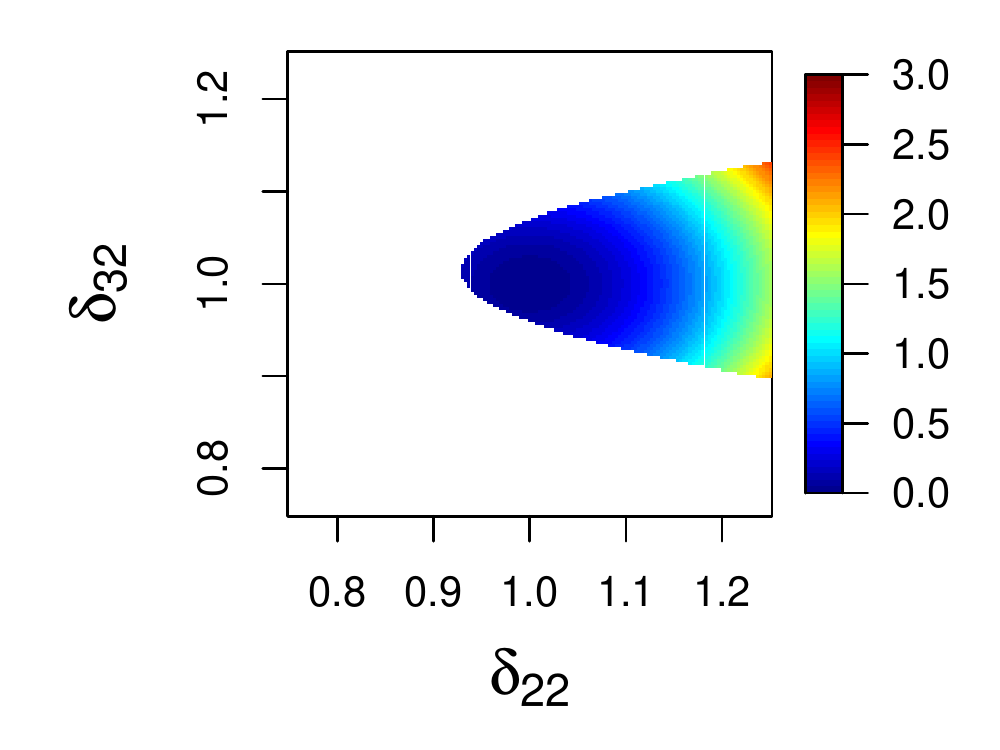}\,\,\,\,\,\,\,\,
\includegraphics[scale=0.45]{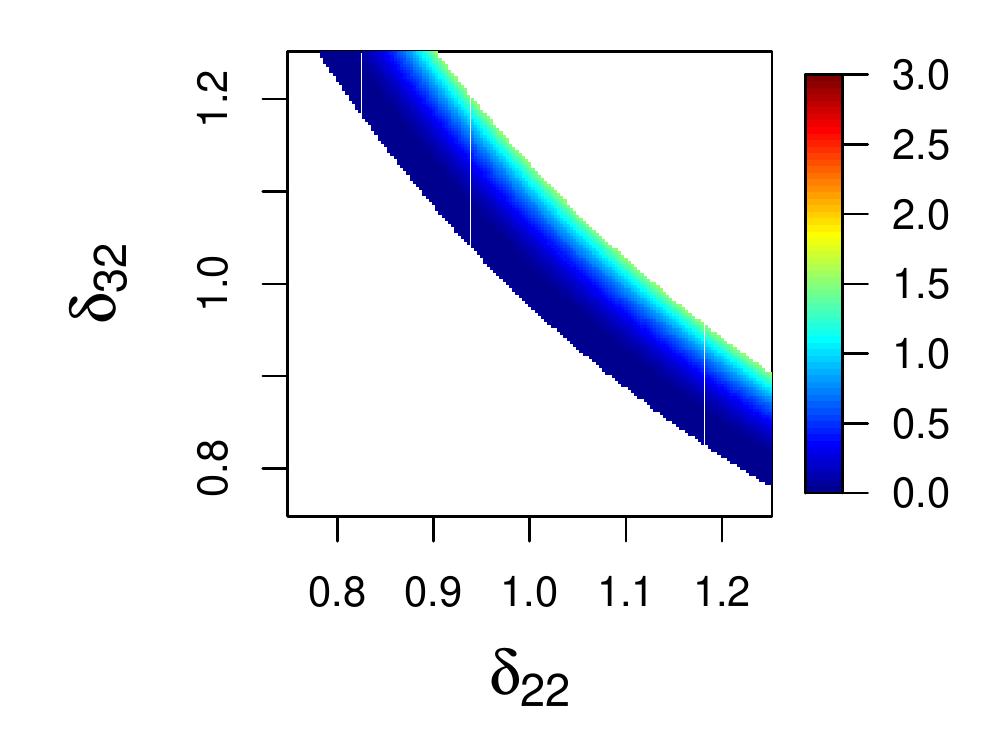}\,\,\,\,\,\,\,
\includegraphics[scale=0.45]{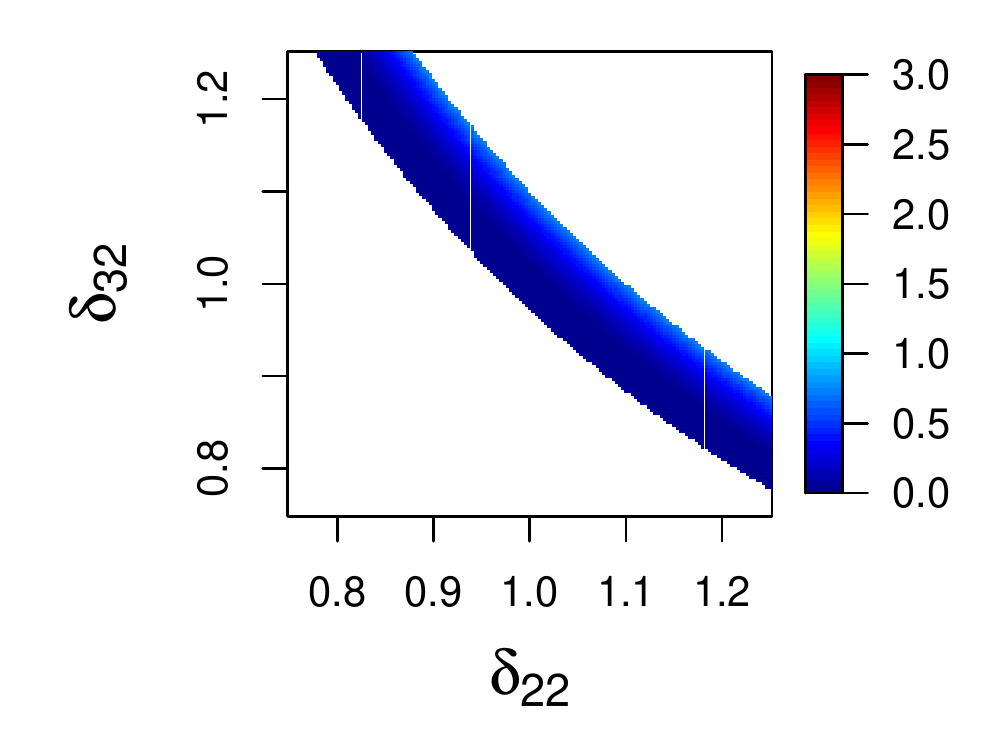}
\end{center}
\caption{Frobenius norm for multi-way variation of the parametes $\sigma_{22}$ and $\sigma_{32}$ of the parameters of the network in Figure \ref{fig:BNex}: standard (left); partial (central); column-based (right).\label{fig:Fro2}}
\end{figure}

In  Figures \ref{fig:KL} and  \ref{fig:Fro} the distance between the original and the varied distribution is not reported for all possible variations since for such variations the resulting covariance matrix is not positive semidefinite. This is even more evident in Figures \ref{fig:KL2} and \ref{fig:Fro2} reporting the KL divergence and the Frobenius norm, respectively, for the multi-way sensitivity analysis of the parameters $\sigma_{22}$ and $\sigma_{33}$. In these plots the white regions correspond to combinations of variations such that the resulting covariance matrix is not positive semidefinite: such a region is very different for the case of the standard sensitivity method (on the left) and the model-preserving ones (on the right and in the center).

Notice that Figures \ref{fig:KL2} and \ref{fig:Fro2}  do not report the divergence between the original and varied distribution in the case of full model-preserving covariations since their inclusion would have made the other plots not particularly informative: the KL divergence for full covariations can be shown to be way smaller than the others  reported in Figure \ref{fig:KL2}, whilst its Frobenius norm is considerably larger. Conversely, standard, partial and column-based (row-based is not included since for this example it would coincide with the partial one) have comparable divergences. However, as expected, the Frobenius norm is smaller for the standard method, although the difference does not appear to be very large.

\subsection{A Real-World Application}
\begin{figure}
\begin{center}
\includegraphics[scale=0.6]{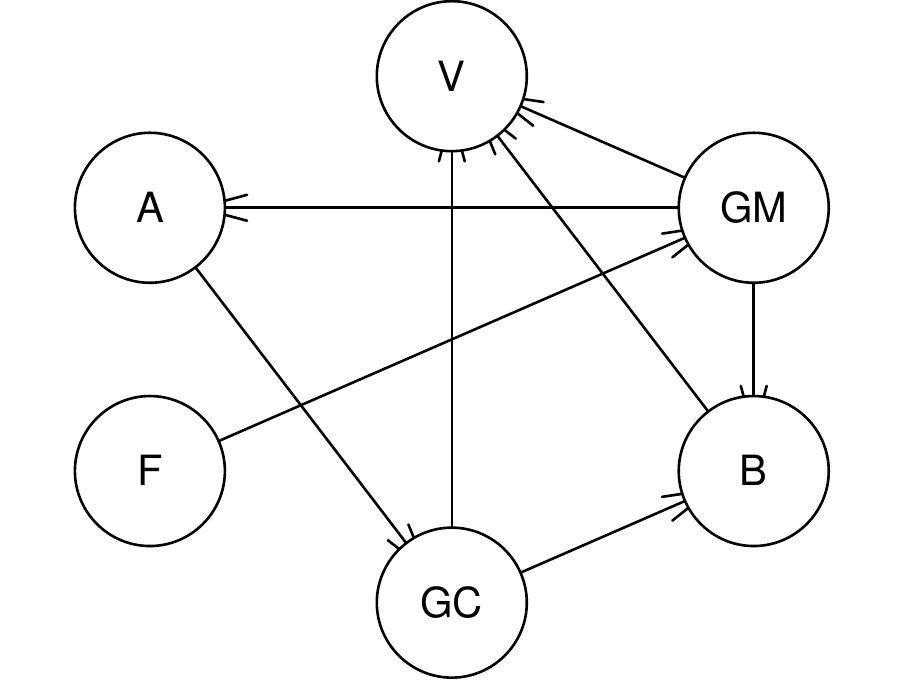}
\includegraphics[scale=0.6]{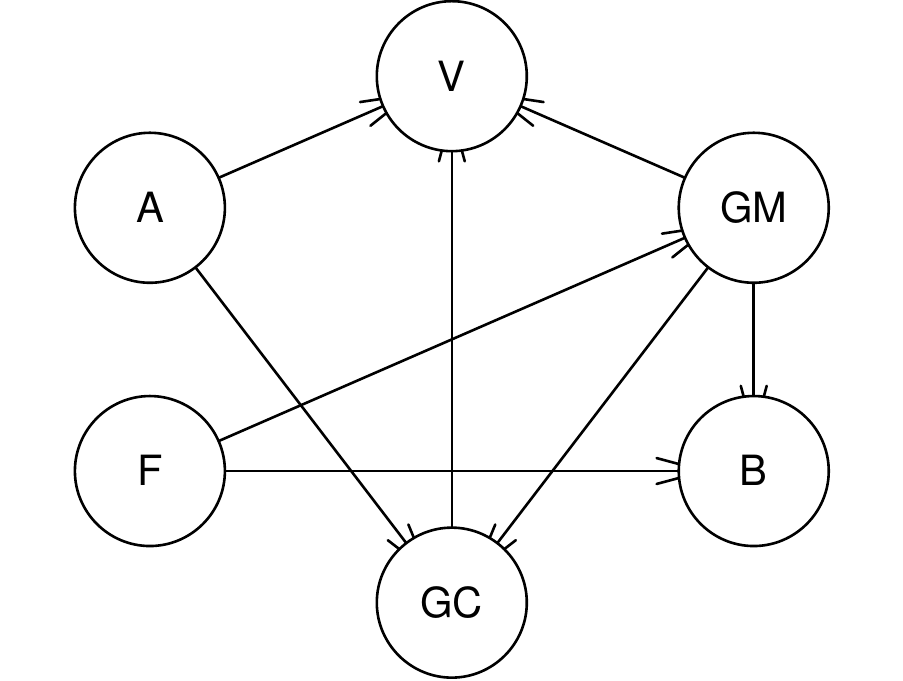}
\end{center}
\caption{Learnt BN model of the metabolics for patients with Cachexia (left network) and for the control group (right network). \label{fig:BNapp}}
\end{figure}

In this section we study a subset of the dataset of \cite{Eisner2011} including metabolomic information of 77 individuals: 47 of them suffering of cachexia, whilst the remaining do not. Cachexia is a metabolic syndrome characterized by loss of muscle with or without loss of fat mass. Although the study of \cite{Eisner2011} included 71 different metabolics which could possibly distinguish individuals who suffer of Cachexia from those who do not, for our illustrative purposes we focus on only six of them: Adipate (A), Betaine (B), Fumarate (F), Glucose (GC), Glutamine (GM) and Valine (V). Two Gaussian BN models were learnt for the two different populations (ill and not ill) using the \texttt{bnlearn} R package \citep{Scutari} resulting in the networks in Figure \ref{fig:BNapp}. The order of the variables was kept fixed for the two populations for ease of comparison. The estimated covariance matrix for individuals suffering of Cachexia is 
\begin{equation}
\begin{blockarray}{ccccccc}
&\text{B} & \text{V} & \text{GC} & \text{GM} & \text{A}&\text{F} \\
\begin{block}{c(cccccc)}
 \text{B}& 304 & 3262 & 220 & 2963 & 414 & 208 \\
  \text{V}& 3262 & 98456 & 6637 & 89431 & 12489 & 6279  \\
  \text{GC}& 220 & 6637 & 3950 & 53223 &  1693 & 839 \\
  \text{GM}& 2963 & 89431 & 53223 & 3050126 & 65012 & 31858 \\
  \text{A}& 414 & 12489 & 1695 & 65012  & 7279 & 1791 \\
   \text{F}& 208 & 6279 &  839 & 31858 & 1791 & 1124 \\
\end{block}
\end{blockarray}
\end{equation}
whilst for the control group this is estimated as
\begin{equation}
\begin{blockarray}{ccccccc}
&\text{B} & \text{V} & \text{GC} & \text{GM} & \text{A}&\text{F} \\
\begin{block}{c(cccccc)}
 \text{B}& 41 & 1004 & 0 &  310 & 168 & 51 \\
  \text{V}& 1004 & 38647 & 0 & 11923 & 10192 & 1974  \\
  \text{GC}& 0 & 0 & 109 & 376 & 0 & 77  \\
  \text{GM}& 310 & 11923 & 376 & 8952 & 3144 & 1092 \\
  \text{A}& 168 & 10192 & 0 & 3144 &  5171 & 520 \\
   \text{F}& 51 & 1974 & 77 & 1092 & 520 & 192 \\
\end{block}
\end{blockarray}~.
\end{equation}

After transforming the two above covariance matrices into correlations it was observed that only two covariances had a disagreement larger than 0.4 (in correlation scale)  between the following Metabolics:  Glutamine/Betaine and Glutamine/Valine. Therefore these are considered of interest. Furthermore there  is interest in the covariance between Glucose/Betaine and Glucose/Valine since these pairs are estimated independent in the control group network, whilst they are dependent for patients suffering of Cachexia. A sensitivity analysis over these parameters is carried out for the network learnt using the data of patients suffering of Cachexia to investigate its robustness. For ease of exposition, we report here a one-way sensitivity analysis over such parameters only, though multi-way analyses can be conducted as formalized in Section \ref{sec:multi} and illustrated in Section \ref{sec:ex1}.

\begin{figure}
\begin{center}
\includegraphics[scale=0.5]{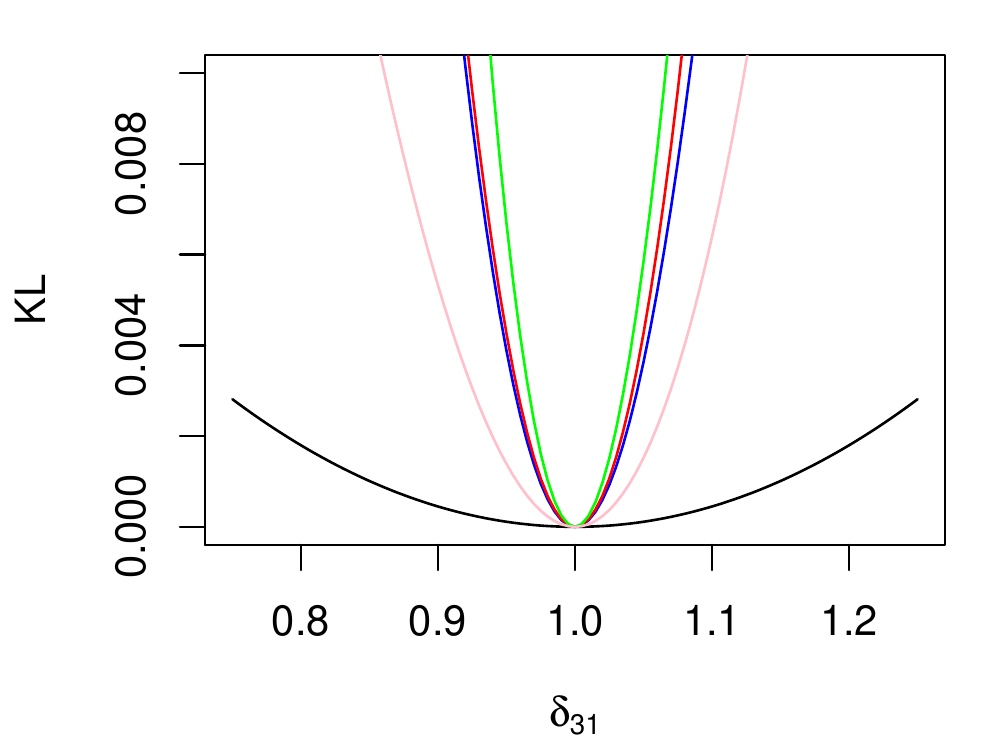}\;\;\;\;
\includegraphics[scale=0.5]{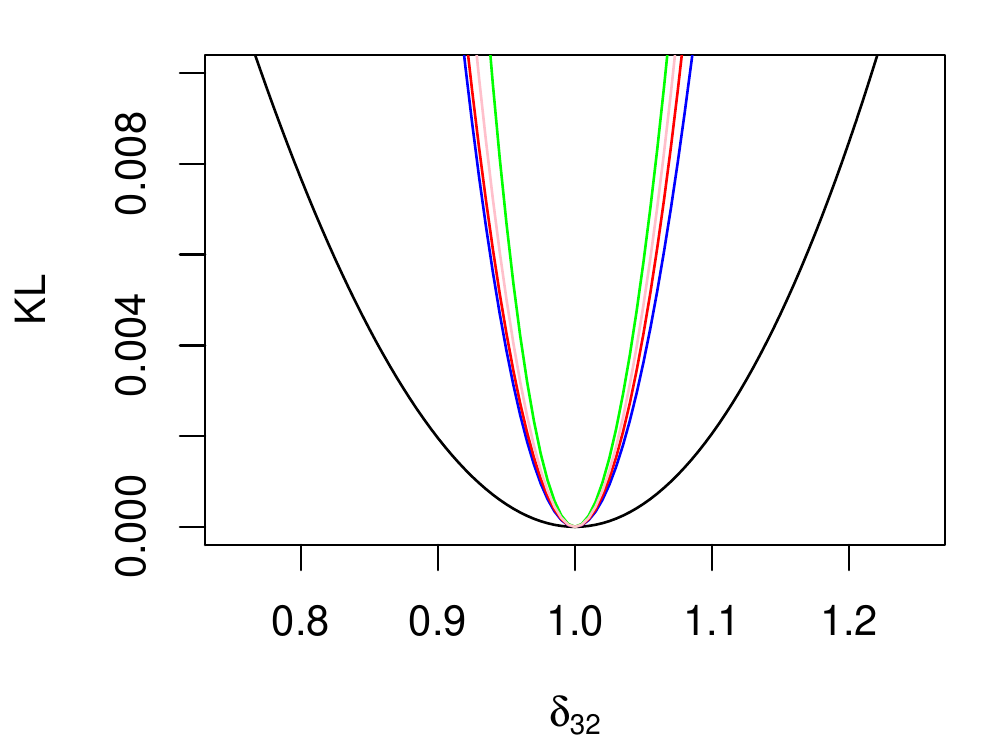}
\includegraphics[scale=0.5]{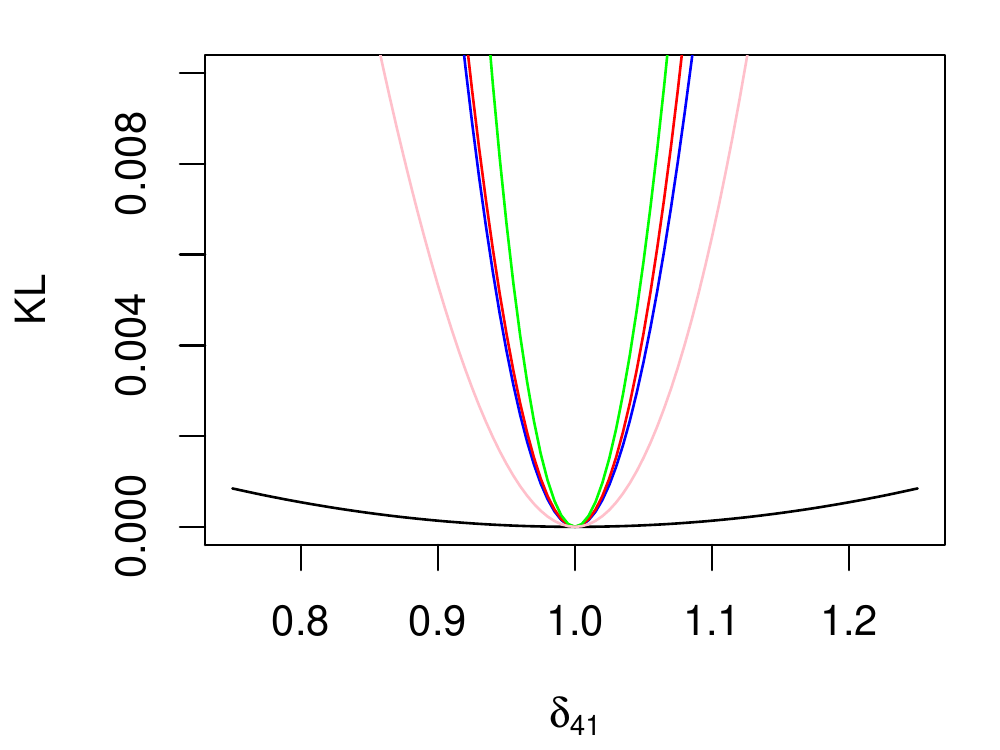}\;\;\;\;
\includegraphics[scale=0.5]{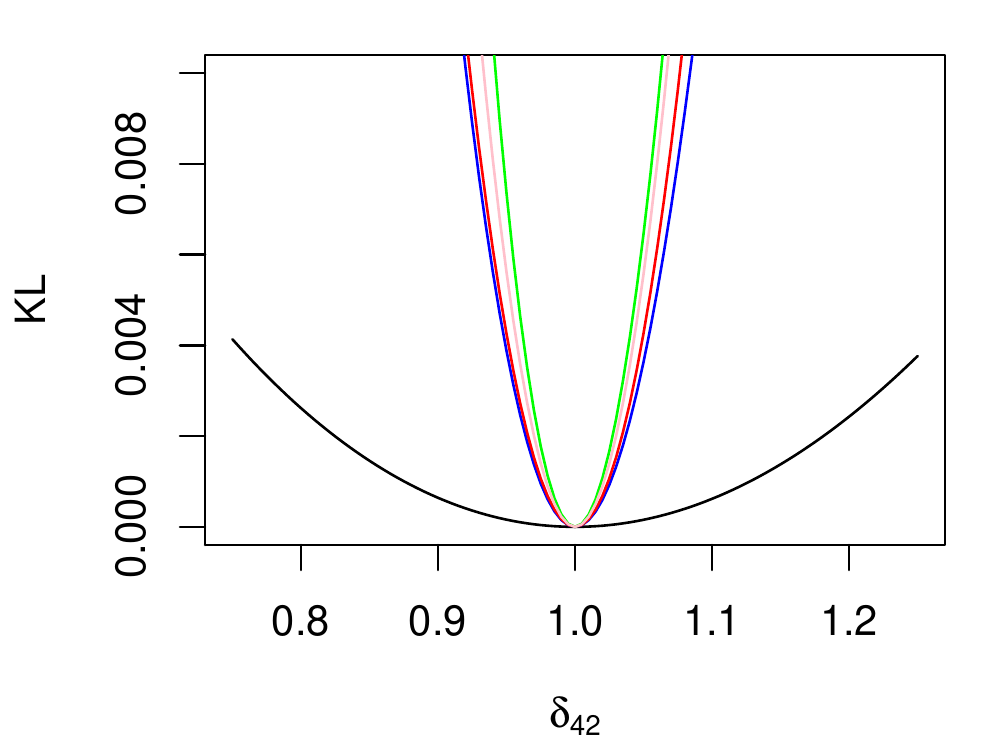}
\end{center}
\caption{KL divergence for one-way variations $\sigma_{ij}\mapsto\delta_{ij}\sigma_{ij}$ of the parameters of the network for patients suffering of Cachexia. We use the color code black = standard variation; blue = full;  red = partial; green = row-based; pink = column-based.\label{fig:KL1}}
\end{figure}

Figure \ref{fig:KL1} reports the KL divergence for the chosen parameters of the network for patients suffering of Cachexia. We can notice that conversely to the sensitivity analysis carried out in Section \ref{sec:ex1}, now standard methods have a much smaller KL divergence than model preserving ones. Furthermore, variations of different parameters lead to substantially different KL divergences under the traditional approach. For model-preserving variations we notice that the KL divergences are fairly similar for variations of different parameters. In addition, row-based model-preserving variations lead to significantly smaller KL divergences in two out of four cases. Thus, based on the result from both row-based and traditional methods, the covariances between Glucose/Valine and Glutamine/Valine appear to have a much stronger effect on the robustness of the network and therefore the validity of their estimated values needs to be carefully validated, for instance using expert information.

\begin{figure}
\begin{center}
\includegraphics[scale=0.5]{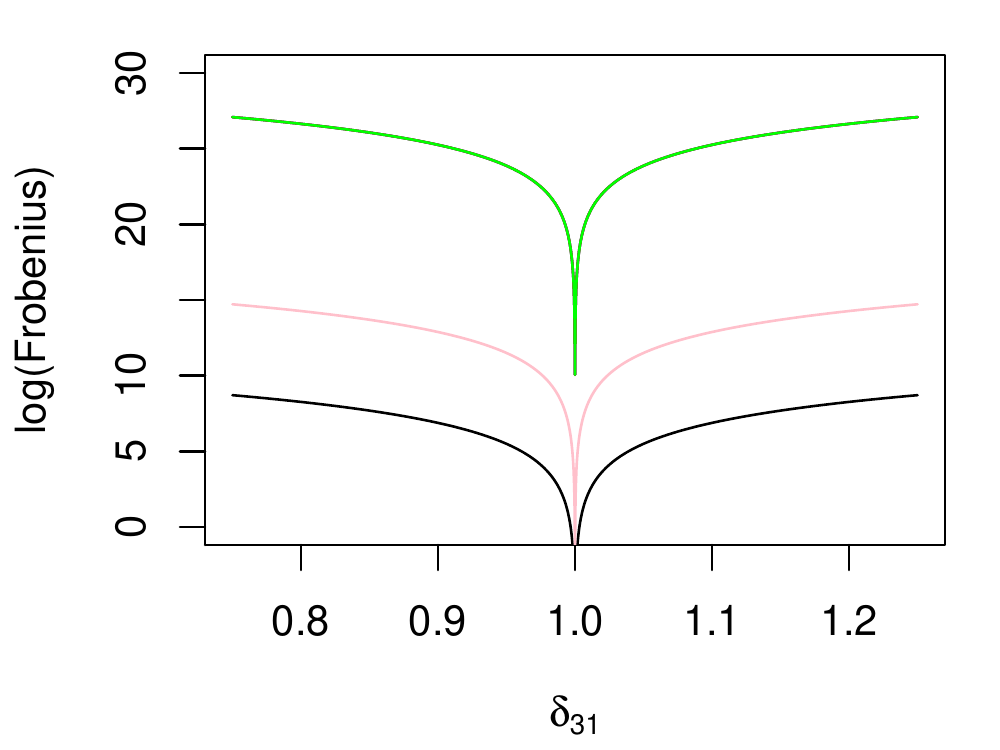}\;\;\;\;
\includegraphics[scale=0.5]{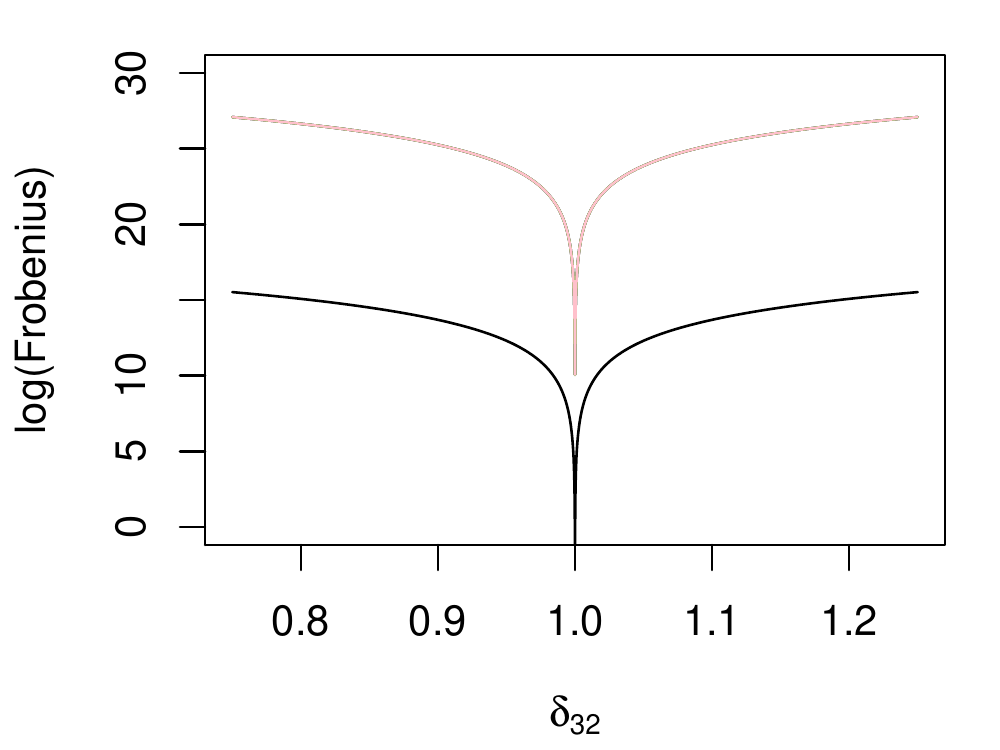}
\includegraphics[scale=0.5]{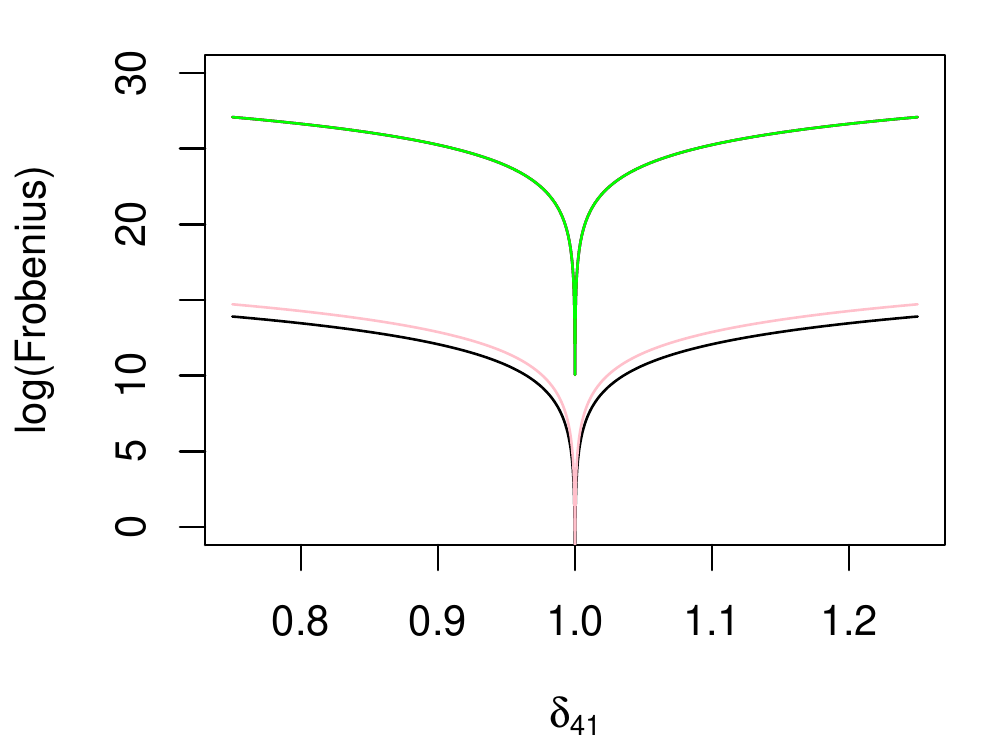}\;\;\;\;
\includegraphics[scale=0.5]{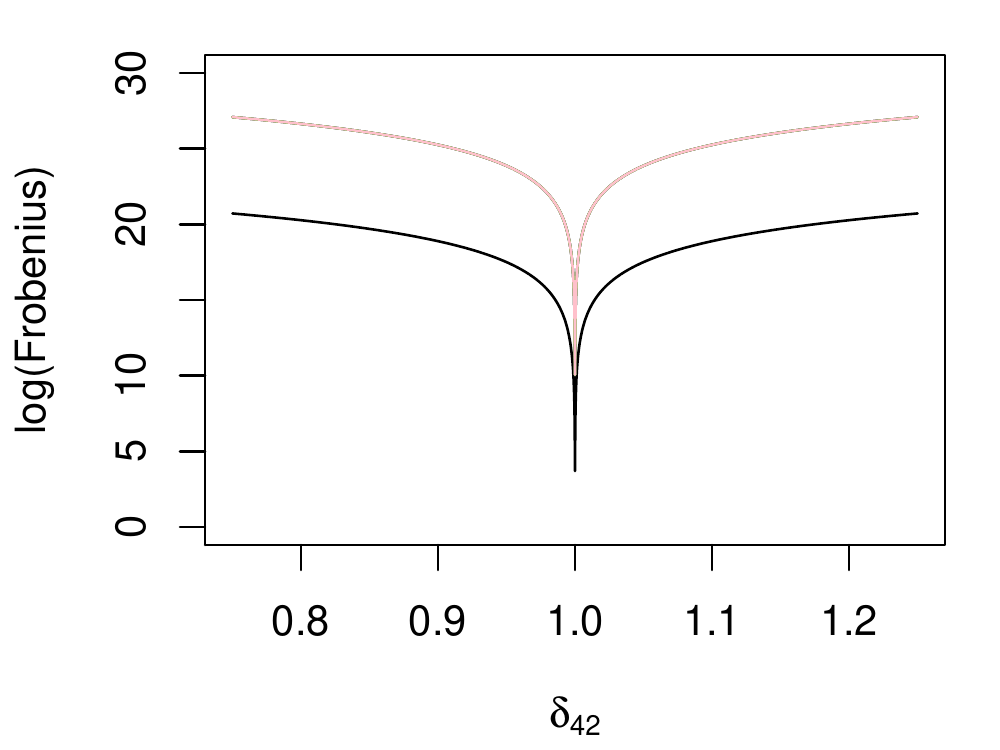}
\end{center}
\caption{KL divergence for one-way variations $\sigma_{ij}\mapsto\delta_{ij}\sigma_{ij}$ of the parameters of the network in Figure \ref{fig:BNex}. We use the color codes black = standard variation; blue = full;  red = partial; green = row-based; pink = column-based.\label{fig:Fro1}}
\end{figure}

Similar conclusions can be drawn from Figure \ref{fig:Fro1} reporting the logarithm of the Frobenius norm for the different parameter variations. For these plots, the goodness of the row-based model-preserving scheme is much more evident and especially for the covariance between Glutamine/Betaine, the Frobenius norm of such scheme is almost equal to the one of the traditional approach. Notice that, because of the structure of the covariance matrix for this example, all variations considered were admissible and lead to a positive semidefinite matrix, irrespective of the approach used. 

\section{Discussion}
Algebraic tools have proved to be extremely powerful to characterize conditional independence models and inferences based on such models. Here we have taken advantage of these tools to  perform sensitivity analyses in Gaussian BNs which do not break the structure of the model. We demonstrated through various examples that our new methods are robust, meaning that the divergences computed under our paradigm are often comparable to those arising from standard methods, with the difference that in our approach the underlying network continues to be a coherent representation of the model.

Here we have assumed that the resulting covariance matrix after a covariation is positive semidefinite. As demonstrated in our examples this sadly is not always the case. Checking whether a matrix is positive semidefinite can now be done almost instantaneously by built-in functions in a variety of software. For standard sensitivity methods, \citet{Gomez2013} used the theory of interval matrices \citep{Rohn1994} to identify allowed variations, those such that the new covariance matrix is positive semidefinite. To our knowledge, there is currently no algebraic theory which characterizes positive semidefinitess for Schur products of matrices. A promising topic of further research is to develop such a theory and utilize it within the context of sensitivity analysis. 

However, we notice that the covariance matrix resulting from a full model-preserving covariation is always positive semidefinite. This was also depicted in Figures \ref{fig:KL} and \ref{fig:Fro}. More formally a symmetric matrix $\Sigma\in\mathbb{R}^{n\times n}$ is positive semidefinite  if and only if $x^{\top}\Sigma x\geq 0$ for all $x\in\mathbb{R}^n$. If $\Sigma$ is multiplied by a positive constant $\delta$, as in full model-preserving analyses, then by default $x^{\top}\delta\Sigma x\geq 0$ for all $x\in\mathbb{R}^n$. Conversely for the other model-preserving covariations which do not vary all entries, positive semidefinitess does not hold as straightforwardly.

We are currently developing a package in the open-source \texttt{R} software \citep{R} to perform sensitivity analysis for both discrete and Gaussian BN models. For the Gaussian case, the aim is to implement both standard and model-preserving methods. Given that currently almost no software allows for sensitivity studies, the development of such a package is critical and could be of great benefit for the whole AI community.

\section*{Acknowledgements}
Christiane G\"orgen and Manuele Leonelli were supported by the programme \enquote{Oberwolfach Leibniz Fellows} of the Mathematisches Forschungsinstitut Oberwolfach in 2017.

\vskip 0.2in
\bibliography{BIB.bib}

\appendix
\section{Proofs}
\subsection{Derivation of equation (\ref{eq:KLknown})}
Plugging the definition of $Y$ and $\tilde{Y}$ into equation (\ref{eq:KL}) we have
\begin{align}
\KL(\tilde{Y}||Y)&=\frac{1}{2}\left(\tr(\Sigma^{-1}(\Delta+\Sigma))+\delta\Sigma^{-1}\delta-n+\ln\left(\frac{\det(\Sigma)}{\det(\Sigma+\Delta)}\right)\right)\\
&=\frac{1}{2}\left(\tr(\Sigma^{-1}\Delta)+\tr(\Sigma^{-1}\Sigma))+\delta\Sigma^{-1}\delta-n+\ln\left(\frac{\det(\Sigma)}{\det(\Sigma+\Delta)}\right)\right)\\
&=\frac{1}{2}\left(\tr(\Sigma^{-1}\Delta)+n+\delta\Sigma^{-1}\delta-n+\ln\left(\frac{\det(\Sigma)}{\det(\Sigma+\Delta)}\right)\right)
\end{align}
and the result follows.
\subsection{Proof of Theorem \ref{theo:total}}
\label{appendix1}
From Proposition \ref{prop:drton}, the result follows if all  $(\#C+1)\times(\#C+1)$ minors of  $(\tilde\Delta\circ\Delta\circ\Sigma)_{A\cup C,B\cup C}$ vanish. First recall that by  Leibniz formula we can write any $(\#C+1)\times(\#C+1)$  minor of $\Sigma_{A\cup C,B\cup C}$ as a polynomial $g$
\begin{equation}\label{eq:minor}
g(\Sigma_{A\cup C,B\cup C})=\sum_{\mathclap{\tau\in S_{\#C+1}}}\sgn(\tau)\prod_{i=1}^{\mathclap{\#C+1}}\sigma_{i\tau(i)},
\end{equation}
where $S_{\#C+1}$ denotes the symmetric group of permutations of the $\#C+1$ indices and $\sgn(\tau)$ is the signature of $\tau$. Since for both total and partial covariation matrices all the entries of $(\tilde\Delta\circ\Delta)_{A\cup C,B\cup C}$ are equal to $\delta$, we have  that
\begin{equation}\label{eq:minor}
g((\tilde\Delta\circ\Delta\circ\Sigma)_{A\cup C,B\cup C})=\sum_{\mathclap{\tau\in S_{\#C+1}}}\sgn(\tau)\prod_{i=1}^{\mathclap{\#C+1}}\delta\sigma_{i\tau(i)}=\delta^{\#C+1}g(\Sigma_{A\cup C,B\cup C})
\end{equation}
which is equal to zero since $g(\Sigma_{A\cup C,B\cup C})=0$ by Lemma~\ref{prop:drton}.

\subsection{Proof of Theorem \ref{theo:1}}
In analogy to the proof of Theorem \ref{theo:total}, the result follows if all $(\#C+1)\times(\#C+1)$ minors of  $(\tilde\Delta\circ\Delta\circ\Sigma)_{A\cup C,B\cup C}$ vanish. To prove the result in this case we use Laplace expansion formula which states that any $(\#C+1)\times(\#C+1)$ minor of $\Sigma_{A\cup C,B\cup C}$ can be written as a polynomial $g$
\begin{equation}
g(\Sigma_{A\cup C,B\cup C})=\sum_{j=1}^{\#C+1}(-1)^{i+j}\sigma_{ij}\det \Sigma_{A\cup C,B\cup C}^{-ij}=\sum_{i=1}^{\#C+1}(-1)^{i+j}\sigma_{ij}\det \Sigma_{A\cup C,B\cup C}^{-ij},
\end{equation}
where $\Sigma_{A\cup C,B\cup C}^{-ij}$ denotes the matrix $\Sigma_{A\cup C,B\cup C}$ without the $i$-th row and the $j$-th column.

Start considering the case $(i,j)$ or $(j,i)\in (A,B)$ and suppose for a row-based covariation $E=\{i\}$. Then
\begin{align}
g((\tilde\Delta\circ\Delta\circ\Sigma)_{A\cup C,B\cup C})&=\sum_{k=1}^{\#C+1}(-1)^{i+k}\delta\sigma_{ik}\det (\tilde\Delta\circ\Delta\circ\Sigma)_{A\cup C,B\cup C}^{-ik}\\
&=\delta\sum_{k=1}^{\#C+1}(-1)^{i+k}\sigma_{ik}\det (\tilde\Delta\circ\Delta\circ\Sigma)_{A\cup C,B\cup C}^{-ik} \label{eq:laplace}
\end{align}
where we use superscripts in matrices to denote rows and columns to be eliminated.

The result follows if $\det (\tilde\Delta\circ\Delta\circ\Sigma)_{A\cup C,B\cup C}^{-ik}=\det\Sigma_{A\cup C,B\cup C}^{-ik}$ for all $k=1,\dots \#C+1$. However this is true since, for $E=\{i\}$, $\delta$s are only in entries $(i,k)$ and  no entries $(k,i)$, that need to be equal to $\delta$ for symmetry, are in $(A\cup C,B\cup C)$.

Consider next the case $E=\{i,l\}$ for a row-based covariation. Then in this case $\det (\tilde\Delta\circ\Delta\circ\Sigma)_{A\cup C,B\cup C}^{-ik}\neq\det\Sigma_{A\cup C,B\cup C}^{-ik}$. However, using again Laplace formula in equation~(\ref{eq:laplace}), we have that
\begin{align}
g((\tilde\Delta\circ\Delta\circ\Sigma)_{A\cup C,B\cup C})&=
\delta\sum_{k=1}^{\#C+1}(-1)^{i+k}\sigma_{ik}\sum_{r=1}^{\#C}(-1)^{l+r}\delta\sigma_{lr}\det(\tilde\Delta\circ\Delta\circ\Sigma)_{A\cup C,B\cup C}^{-\{i,l\}\{k,r\}}\nonumber\\
&= \delta^2\sum_{k=1}^{\#C+1}(-1)^{i+k}\sigma_{ik}\sum_{r=1}^{\#C}(-1)^{l+r}\sigma_{lr}\det(\tilde\Delta\circ\Delta\circ\Sigma)_{A\cup C,B\cup C}^{-\{i,l\}\{k,r\}}.
\end{align}
The result follows again since $\det(\tilde\Delta\circ\Delta\circ\Sigma)_{A\cup C,B\cup C}^{-\{i,l\}\{k,r\}}=\det(\Sigma)_{A\cup C,B\cup C}^{-\{i,l\}\{k,r\}}$. Laplace expansion can now be used iteratively to demonstrate that all row-based covariation matrices with $E\subseteq A$ induce a model-preserving map when $(i,j)$ or $(j,i)\in (A,B)$. 

The result follows using the same reasoning for column-based covariation matrices when $(i,j)$ or $(j,i)\in (A,B)$ by using the Laplace formula expansion over the rows of the matrix. The result is equally proven for row-based covariations when $(i,j)$ or $(j,i)\in (A,C)$ and column-based covariations when $(i,j)$ or $(j,i)\in (C,B)$.

The proof of the result needs to be slightly adapted when $\delta$s appear in the submatrix $(\tilde\Delta\circ\Delta)_{C,C}$: that is for column-based covariation if $(i,j)$ or $(j,i)\in (A,C)$, for row-based covariation if $(i,j)$ or $(j,i)\in (C,B)$ and for both covariations if $(i,j)$ and $(j,i)\in (C,C)$ where for symmetric reasons both belong to the submatrix. In such cases, because the matrix $\lfloor(\tilde\Delta\circ\Delta\circ\Sigma)_{A\cup C,B\cup C}\rfloor^1$ needs to be symmetric, extra $\delta$s already appear within $\tilde\Delta_{A\cup C,B\cup C}$. So for instance, if all the entries in the $i$-th row of $(\tilde\Delta\circ\Delta)_{C,C}$ are $\delta$, then also its $i$-th column must have $\delta$s for symmetry. But because of this then we only have that  $\det(\tilde\Delta\circ\Delta\circ\Sigma)_{A\cup C,B\cup C}^{-CC}=\det(\Sigma)_{A\cup C,B\cup C}^{-CC}$, thus requiring us to apply the Laplace expansion over all rows or all columns with index in $C$.

\subsection{Proof of Theorem \ref{theo:compositions}}
Let $\A$ be a parameter set. If $D$ and $D'$ are matrices such that $\Phi_D$ and $\Phi_{D'}$ map $\A$ to a subset of itself then also the composition of these two maps sends $\A$ to a subset of itself, $\Phi_D(\Phi_{D'}(\A))\subseteq\A$. Furthermore, $\Phi_D(\Phi_{D'}(\Sigma))=\Phi_D(D'\circ\Sigma)=D\circ D'\circ\Sigma$ for any matrix $\Sigma$.

\subsection{Derivation of equation (\ref{equazione})}
First notice for a total covariation matrix $\tilde{\Delta}$ that $\tilde{\Delta}\circ\Delta\circ\Sigma=\delta\Sigma$. Thus 
\begin{equation}
\label{1}
\tr(\Sigma^{-1}(\tilde{\Delta}\circ\Delta\circ\Sigma))=\tr(\Sigma^{-1}(\delta\Sigma))=\tr(\delta\Sigma^{-1}\Sigma)=n\delta,
\end{equation}
and
\begin{equation}
\label{2}
\log\left(\frac{\det(\Sigma)}{\det(\tilde{\Delta}\circ\Delta\circ\Sigma)}\right)=\log\left(\frac{\det(\Sigma)}{\det(\delta\Sigma)}\right)=\log\left(\frac{\det(\Sigma)}{\delta^n\det(\Sigma)}\right)=-n\log(\delta).
\end{equation}
Plugging equations (\ref{1}) and (\ref{2}) into equation (\ref{eq:KLour}), we find
\begin{equation}
\KL(\tilde{Y}||Y)=\frac{1}{2}(n\delta-n-n\log(\delta)),
\end{equation}
and the result follows.
\end{document}